%% file: main.tex
\documentclass[10pt,twocolumn,letterpaper]{article}

\usepackage{cvpr}
\usepackage{times}
\usepackage{epsfig}
\usepackage{graphicx}
\usepackage{amsmath}
\usepackage{amssymb}
\usepackage{color}
\usepackage{comment}
\usepackage{subfig}
\usepackage{float}

\usepackage{enumitem}
\setlist{nosep}

\usepackage{wrapfig}
\usepackage{stackengine}
\usepackage[dvipsnames,svgnames,x11names]{xcolor}
\usepackage{rotating}
\usepackage[export]{adjustbox}
\usepackage{xspace}
\makeatletter
\usepackage{caption}
\newcommand{\mycaption}[2]{\caption{\textbf{#1.}\xspace#2}}

\definecolor{darkgreen}{RGB}{0,127,0}

\newif\ifdrafting
\draftingtrue 
\draftingfalse 
\ifdrafting
	\newcommand{\hs}[1]{{\color{magenta}[Hang: #1]}}
	\newcommand{\OG} [1] {\textcolor{darkgreen}{[OG: #1]}}
    \newcommand{\JK}[1]{{\color{Orange}[JK: #1]}}
    \newcommand{\DS}[1]{{\color{red}[DS: #1]}}
    
\else
	\newcommand{\hs}[1]{}
	\newcommand{\OG} [1] {}
	\newcommand{\DS} [1] {}
	\newcommand{\JK} [1] {}
\fi

\def\ie{i.e., }
\def\eg{e.g., }
\def\supp{appendix\xspace}

\usepackage{xcolor}
\usepackage{rotating}
\usepackage[export]{adjustbox}
\makeatletter

\usepackage{booktabs}
\usepackage{array, makecell, tabularx}
\usepackage{multirow}
\usepackage{authblk}
\usepackage[hang,flushmargin]{footmisc}

\usepackage[pagebackref=true,breaklinks=true,colorlinks,bookmarks=false]{hyperref}

\cvprfinalcopy 

\def\cvprPaperID{89} 

\ifcvprfinal\fi
\begin{document}

\title{Pixel-Adaptive Convolutional Neural Networks}
\author[1]{Hang Su}
\author[2]{Varun Jampani}
\author[2]{Deqing Sun}
\author[2]{Orazio Gallo}
\author[1]{Erik Learned-Miller}
\author[2]{Jan Kautz}
\makeatletter 
\renewcommand\AB@affilsepx{\quad \protect\Affilfont} 
\makeatother
\affil[1]{UMass Amherst}
\affil[2]{NVIDIA}

\maketitle

\input{abstract.tex}

\input{intro.tex}

\input{related.tex}

\input{pac.tex}

\input{upsample.tex}

\input{crf.tex}

\input{hotswap.tex}

\input{conclusion.tex}

\input{acknowledgements.tex}

{\small
\bibliographystyle{ieee}
\bibliography{main}
}

\newpage
\clearpage
\appendix

{\raggedleft{} \bf \Large Appendix}
\vspace{0.3cm}

\renewcommand\thesection{\Alph{section}}

In the appendix, we provide additional details and results on the deep joint upsampling experiments (Sec.~\ref{sec:supp_upsample}) and PAC-CRF (Sec.~\ref{sec:supp_crf}). 

\input{supp.tex}

\end{document}

%% file: abstract.tex
\begin{abstract}



Convolutions are the fundamental building blocks of CNNs.
The fact that their weights are spatially shared is one of the main reasons for their widespread use, but it is also a major limitation, as it makes convolutions content-agnostic.
We propose a pixel-adaptive convolution (PAC) operation, a simple yet effective modification of standard convolutions, in which the filter weights are multiplied with a spatially varying kernel that depends on learnable, local pixel features. PAC is a generalization of several popular filtering
techniques and thus can be used for a wide range of
use cases. Specifically, we demonstrate state-of-the-art performance when PAC is used for deep joint image upsampling.
PAC also offers an effective
alternative to fully-connected CRF (Full-CRF), called PAC-CRF, 
which performs competitively compared to Full-CRF, while being considerably 
faster. In addition, we also demonstrate that PAC can be used as a drop-in replacement for convolution layers in pre-trained networks, resulting in consistent performance improvements.

\end{abstract}

%% file: intro.tex
\vspace{-3mm}
\section{Introduction}
\label{sec:intro}
\vspace{-2mm}


Convolution is a basic operation in many image processing 
and computer vision applications and the major building 
block of Convolutional Neural Network (CNN) architectures. 
It forms one of the most prominent ways of 
propagating and integrating features across image pixels due to its 
simplicity and highly optimized CPU/GPU implementations. 
In this work, we concentrate on two important characteristics of standard
spatial convolution and aim to alleviate some of its drawbacks:
\emph{Spatial Sharing} and its \emph{Content-Agnostic} nature. 

\emph{Spatial Sharing}: A typical CNN shares 
filters' parameters across the whole input. In addition to affording translation invariance to the CNN, spatially invariant 
convolutions significantly reduce the number of parameters compared with fully connected layers.
However, spatial sharing is not without drawbacks.
For dense pixel prediction tasks, such as semantic segmentation,
the loss is spatially varying because of varying scene elements on a pixel grid.
Thus the optimal gradient direction for parameters differs at each pixel.
However, due to the spatial sharing nature of convolution, the loss gradients from
all image locations are globally pooled to train each filter.
This forces the CNN to learn filters that minimize the error across all pixel locations at once, but may be sub-optimal at any specific location.


\begin{figure}
\begin{center}
  \centerline{\includegraphics[width=0.9\columnwidth]{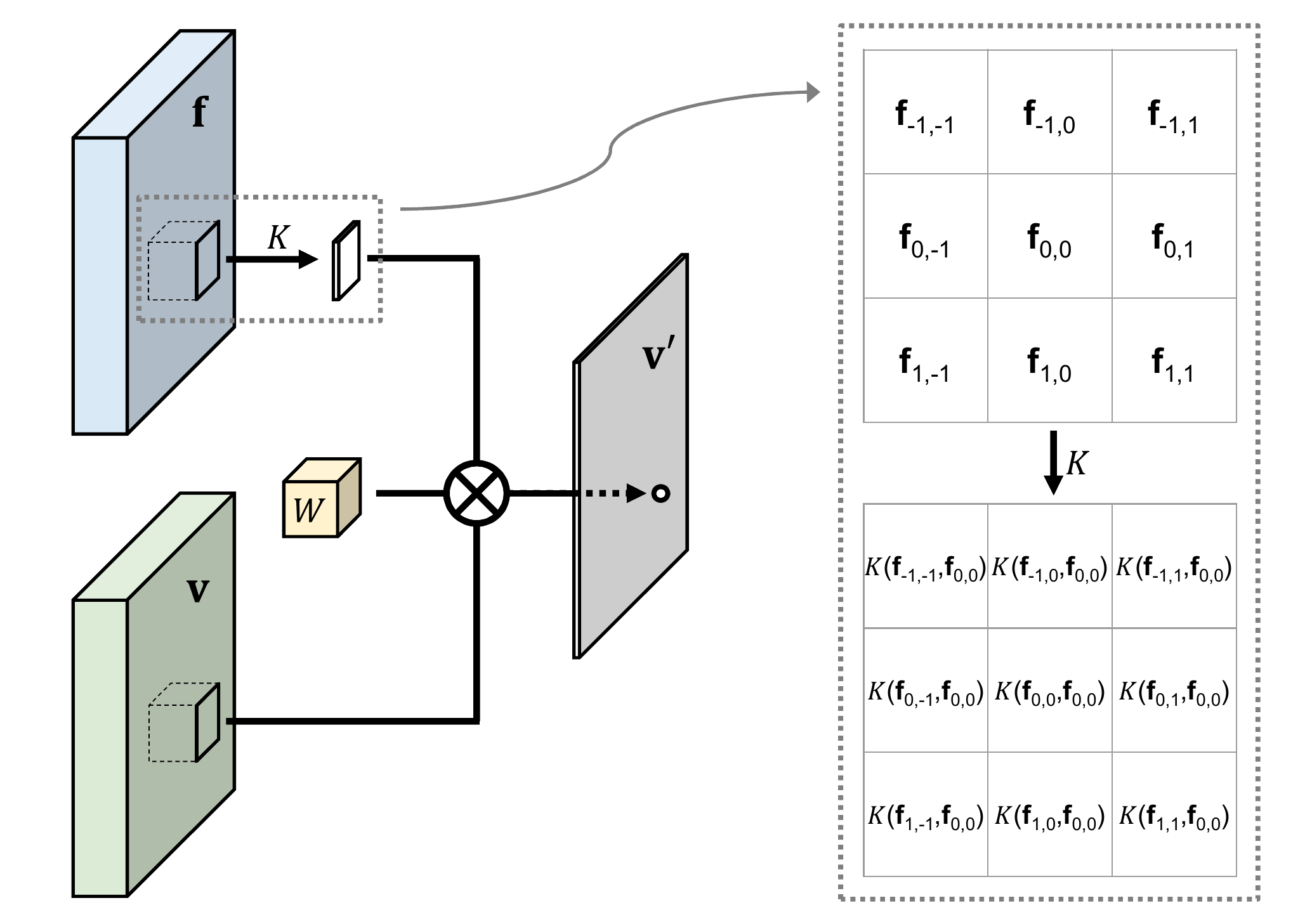}} \vspace{-0.0cm} 
  \mycaption{Pixel-Adaptive Convolution}{PAC modifies a standard
  convolution on an input $\mathbf{v}$ by modifying the spatially invariant filter
  $\mathbf{W}$ with an adapting kernel $K$. The adapting kernel is constructed using either
  pre-defined or learned features $\mathbf{f}$.
  \(\otimes\) denotes element-wise multiplication of matrices followed by a summation. Only one output channel is shown for the illustration. }
  \label{fig:pac}
\end{center}
\end{figure}

\emph{Content-Agnostic}: Once a CNN is trained, the same convolutional 
filter banks are applied to all the images and all the pixels
irrespective of their content. The image content varies
substantially across images and pixels.
Thus a single trained CNN may not be
optimal for all image types (\eg images taken in daylight
and at night) as well as different pixels in an image (\eg sky vs.\ pedestrian pixels). 
Ideally, we would like CNN filters to be adaptive
to the type of image content, which is not the case with standard CNNs.
These drawbacks can be tackled by 
learning a large number of filters in an attempt to capture both image and pixel variations. This, however, increases the number of parameters, requiring a larger memory footprint  
and an extensive amount of labeled data. 
A different approach is to use
\emph{content-adaptive} filters inside the networks.

Existing content-adaptive convolutional networks can be broadly
categorized into two types. One class of techniques make 
traditional image-adaptive filters, such as bilateral filters~\cite{aurich1995non,tomasi1998bilateral} 
and guided image filters~\cite{he2013guided} differentiable, and use them as
layers  inside a CNN~\cite{krahenbuhl2011efficient,li2014meanfield,zheng2015conditional,chen2015learning,chen2016semantic,jampani2016learning,lin2016efficient,chandra2016fast,gadde2016superpixel,liu2017learning,Wang_nonlocalCVPR2018,wu2018fast}.
These content-adaptive layers are usually 
designed for enhancing CNN results but not as a replacement for standard convolutions.
Another class of content-adaptive networks involve learning 
position-specific kernels using a separate sub-network that predicts
convolutional filter weights at each pixel.
These are called ``Dynamic Filter Networks'' (DFN)
~\cite{xue2016visual,jia2016dynamic,dai2017deformable,wu2018dynamic} 
(also referred to as cross-convolution~\cite{xue2016visual} or kernel prediction networks~\cite{bako2017kernel})
and have been shown to be useful in several computer vision tasks. Although
DFNs are generic and can be used as a replacement to standard
convolution layers, such a kernel prediction strategy is difficult to scale
to an entire network with a large number of filter banks.

In this work, we propose a new content-adaptive convolution layer
that addresses some of the limitations of the existing content-adaptive
layers while retaining several favorable properties of spatially
invariant convolution. Fig.~\ref{fig:pac} illustrates our
content-adaptive convolution operation, which we call
``Pixel-Adaptive Convolution" (PAC).
Unlike a typical DFN, where different kernels are
predicted at different pixel locations, we \emph{adapt} a standard 
spatially invariant convolution filter $\mathbf{W}$ at each pixel
by multiplying it with a spatially varying filter $K$,
which we refer to as an ``adapting kernel''.
This adapting kernel has a pre-defined form (e.g., Gaussian 
or Laplacian) 
and depends on the pixel features. For instance, the adapting kernel that we mainly
use in this work is Gaussian: $e^{-\frac{1}{2}||\mathbf{f}_i - \mathbf{f}_j||^2}$,
where $\mathbf{f}_i \in \mathbb{R}^d$ is a $d$-dimensional 
feature at the $i^{th}$ pixel. We refer to these pixel features $\mathbf{f}$ 
as ``adapting features", and they can be either pre-defined, such as pixel position
and color features, or can be learned using a CNN.

We observe that PAC, despite being a simple modification to standard 
convolution, is highly flexible and can be seen as 
a generalization of several widely-used filters.
Specifically, we show that PAC is a generalization of spatial convolution,
bilateral filtering~\cite{aurich1995non,tomasi1998bilateral}, and pooling operations
such as average pooling and detail-preserving pooling~\cite{saeedan2018detail}.
We also implement a variant of PAC that does pixel-adaptive transposed convolution
(also called deconvolution)
which can be used for learnable guided upsampling of intermediate CNN representations.
We discuss more about these generalizations and variants in 
Sec.~\ref{sec:methods}. 

As a result of its simplicity and being a generalization of several
widely used filtering techniques, PAC can be useful in a wide range
of computer vision problems. In this work, we demonstrate its applicability
in three different vision problems. 
In Sec.~\ref{sec:upsampling}, we
use PAC in joint image upsampling networks and obtain state-of-the-art
results on both depth and optical flow upsampling tasks. In 
Sec.~\ref{sec:crf}, we use PAC in a learnable conditional
random field (CRF) framework and
observe consistent improvements with respect to the widely used 
fully-connected CRF~\cite{krahenbuhl2011efficient}. 
In Sec.~\ref{sec:hotswap}, we demonstrate
how to use PAC as a drop-in replacement of 
trained convolution layers in a CNN
and obtain performance improvements after fine-tuning.
In summary, we observe that PAC is highly versatile and has 
wide applicability in a range of computer vision tasks.

%% file: related.tex

\section{Related Work}
\label{sec:related}
\vspace{-3mm}
\vspace{1mm}
\noindent \textbf{Image-adaptive filtering.}
Some important image-adaptive filtering techniques include
bilateral filtering~\cite{aurich1995non,tomasi1998bilateral}, guided image filtering~\cite{he2013guided},
non-local means~\cite{buades2005non,awate2005higher}, and
propagated image filtering~\cite{rick2015propagated}, to name
a few. A common line of research is to make these
filters differentiable and use them as content-adaptive
CNN layers.
Early work~\cite{zheng2015conditional,chen2015learning} 
in this direction back-propagates through bilateral
filtering and can thus leverage fully-connected CRF inference~\cite{krahenbuhl2011efficient} 
on the output of CNNs.
The work of \cite{jampani2016learning} and \cite{gadde2016superpixel} proposes to use bilateral filtering layers inside CNN architectures.~Chandra \etal~\cite{chandra2016fast} propose a layer that performs closed-form Gaussian CRF inference in a CNN.
Chen \etal~\cite{chen2016semantic} and Liu \etal~\cite{liu2017learning} propose differentiable local propagation modules that have roots in domain transform
filtering~\cite{gastal2011domain}.
Wu~\etal~\cite{wu2018fast} 
and Wang~\etal~\cite{Wang_nonlocalCVPR2018} propose neural network layers
to perform guided filtering~\cite{he2013guided} and non-local
means~\cite{Wang_nonlocalCVPR2018} respectively inside CNNs.
Since these techniques are tailored towards a particular CRF
or adaptive filtering technique, they are used for specific
tasks and cannot be directly used as a replacement of general convolution.
Closest to our work are the sparse, high-dimensional neural
networks~\cite{jampani2016learning} which generalize standard 2D convolutions to high-dimensional convolutions, enabling them to
be content-adaptive. Although conceptually more generic than PAC, such high-dimensional networks can not learn the adapting features
and have a larger computational overhead due to the use of specialized lattices and hash tables. 

\vspace{1mm}
\noindent \textbf{Dynamic filter networks.} Introduced by Jia~\etal~\cite{jia2016dynamic}, dynamic filter networks (DFN) are an example of another class of content-adaptive filtering techniques. Filter weights are themselves directly predicted by a separate network branch, and provide custom filters specific to different input data. The work is later extended by Wu \etal~\cite{wu2018dynamic} with an additional attention mechanism and a dynamic sampling strategy to allow the position-specific kernels to also learn from multiple neighboring regions. Similar ideas have been applied to several task-specific use cases, \eg motion prediction~\cite{xue2016visual}, semantic segmentation~\cite{harley2017segmentation}, and Monte Carlo rendering denoising~\cite{bako2017kernel}. Explicitly predicting all position-specific filter weights requires a large number of parameters, so DFNs typically require a sensible architecture design and are difficult to scale to multiple dynamic-filter layers. Our approach differs in that PAC reuses spatial filters just as standard convolution, and only modifies the filters in a position-specific fashion. Dai \etal propose deformable convolution~\cite{dai2017deformable}, which can also produce position-specific modifications to the filters. Different from PAC, the modifications there are represented as \emph{offsets} with an emphasis on learning geometric-invariant features. 

\vspace{1mm}
\noindent \textbf{Self-attention mechanism.} Our work is also related to the self-attention mechanism originally proposed by Vaswani \etal for machine translation~\cite{vaswani2017attention}. Self-attention modules compute the responses at each position while attending to the global context. Thanks to the
use of global information, self-attention has been successfully used in several applications, including image generation~\cite{zhang2018self, parmar2018image} and video activity recognition~\cite{Wang_nonlocalCVPR2018}. Attending to the whole image can be computationally expensive, and, as a result, can only be afforded on low-dimensional feature maps, \eg as in ~\cite{Wang_nonlocalCVPR2018}. Our layer produces responses that are sensitive to a more local context (which can be alleviated through \emph{dilation}), and is therefore much more efficient. 

%% file: pac.tex
\section{Pixel-Adaptive Convolution}
\label{sec:methods}
\vspace{-2mm}
In this section, we start with a formal definition of standard
spatial convolution and then explain our generalization of it
to arrive at our pixel-adaptive convolution (PAC). Later, we
will discuss several variants of PAC and how they are
connected to different image filtering techniques.
Formally, a spatial convolution of image features $\mathbf{v}=(\mathbf{v}_1, \dots, \mathbf{v}_n), \mathbf{v}_i \in \mathbb{R}^{c}$ over $n$ pixels and $c$ channels
with filter weights $\mathbf{W}\in \mathbb{R}^{c' \times c \times s \times s}$ can be written as 
\begin{align}
\mathbf{v}'_i = \sum_{j \in \Omega(i)} \mathbf{W}\left[\mathbf{p}_i - \mathbf{p}_j\right] \mathbf{v}_j + \mathbf{b} 
\label{eq:conv}
\end{align}
where $\mathbf{p}_i=(x_i,y_i)^\intercal$ are pixel coordinates, $\Omega(\cdot)$ defines an $s\times s$ convolution window, and $\mathbf{b}\in \mathbb{R}^{c'}$ denotes biases. With a slight abuse of notation, we use $[\mathbf{p}_i - \mathbf{p}_j]$ to denote indexing of the spatial dimensions of an array with 2D spatial offsets. 
This convolution operation results in a $c'$-channel output,
$\mathbf{v}'_i \in \mathbb{R}^{c'}$, at each pixel $i$.
\DS{not sure whether we have time to make it consistent: why not $\mathbf{W}$, i.e., boldface for both vector and matrix? Note that we use    $K(\mathbf{f}_i,\mathbf{f}_j)=1,\; W=\frac{1}{F^2}\cdot \mathbf{1}$ to explain pooling. Note the $\mathbf{1}$. }
%
Eq.~\ref{eq:conv} highlights how the weights only depend on pixel position and thus are agnostic to image content. 
%
%
In other words, the weights are \emph{spatially shared} and, therefore, \emph{image-agnostic}.
As outlined in Sec.~\ref{sec:intro}, these
properties of spatial convolutions are limiting: we
would like the filter weights $\mathbf{W}$ to be content-adaptive.

One approach to make the convolution operation content-adaptive, rather than only based on pixel locations, is to 
generalize $\mathbf{W}$ to depend on the pixel features, $\mathbf{f} \in \mathbb{R}^d$:
\begin{align}
\mathbf{v}'_i = \sum_{j \in \Omega(i)} \mathbf{W}\left(\mathbf{f}_i - \mathbf{f}_j\right) \mathbf{v}_j + \mathbf{b}
\label{eq:nd-conv}
\end{align}
where $\mathbf{W}$ can be seen as a high-dimensional filter operating
in a $d$-dimensional feature space.
In other words, we can apply Eq.~\ref{eq:nd-conv} by
first projecting the input signal $\mathbf{v}$ into a
$d$-dimensional space, and then performing $d$-dimensional
convolution with $\mathbf{W}$.
Traditionally, such high-dimensional filtering is limited
to hand-specified filters such as Gaussian filters~\cite{adams2010fast}. 
Recent work~\cite{jampani2016learning} lifts this restriction and proposes a
technique to freely parameterize and learn $\mathbf{W}$ in
high-dimensional space. Although generic and
used successfully in several computer vision 
applications~\cite{jampani2016learning,jampani2017video,su2018splatnet}, 
high-dimensional convolutions have several shortcomings. 
First, since 
data projected on a higher-dimensional space is sparse,
special lattice structures and hash tables are needed to perform the
convolution~\cite{adams2010fast} resulting in considerable
computational overhead. Second, it is difficult to learn
features $\mathbf{f}$ resulting in the use of hand-specified
feature spaces such as position and color features,
$\mathbf{f} = (x,y,r,g,b)$. Third, we have to restrict 
the dimensionality $d$ of features (say, $<10$) as the
projected input image can become too sparse in high-dimensional spaces. In addition, the advantages that
come with spatial sharing of standard convolution are lost with
high-dimensional filtering.
\DS{Remember to add citations}

\vspace{1mm}
\noindent \textbf{Pixel-adaptive convolution.}
Instead of bringing convolution to higher dimensions, which
has the above-mentioned drawbacks, we choose to modify the
spatially invariant convolution in Eq.~\ref{eq:conv} with a
spatially varying kernel $K \in \mathbb{R}^{c' \times c \times s 
\times s}$ that depends on pixel features $\mathbf{f}$:
\begin{equation}
\mathbf{v}'_i = \sum_{j \in \Omega(i)} K\left(\mathbf{f}_i, \mathbf{f}_j\right) \mathbf{W}\left[\mathbf{p}_i - \mathbf{p}_j\right] \mathbf{v}_j + \mathbf{b} \label{eq:pac}
\end{equation}
where $K$ is a kernel function that has a fixed parametric form
such as Gaussian: $K(\mathbf{f}_i, \mathbf{f}_j)=\exp 
(-\frac{1}{2}(\mathbf{f}_i-\mathbf{f}_j)^\intercal 
(\mathbf{f}_i-\mathbf{f}_j))$. \DS{the Gaussian kernel has been defined in intro in a more succinct way $e^{-|\mathbf{f}_i - \mathbf{f}_j|^2}$; they differ by 1/2.} Since $K$ has a pre-defined form and is not parameterized
as a high-dimensional filter, we can perform this
filtering on the 2D grid itself without moving onto higher
dimensions.
We call the above filtering operation (Eq.~\ref{eq:pac}) as
``Pixel-Adaptive Convolution" (PAC) because the standard spatial
convolution $\mathbf{W}$ is adapted at each pixel using pixel features
$\mathbf{f}$ via kernel $K$. We call these pixel features
$\mathbf{f}$ as ``adapting features" and the kernel $K$ as
``adapting kernel". The adapting features $\mathbf{f}$
can be either hand-specified such as position and color
features $\mathbf{f} = (x,y,r,g,b)$ or can be deep
features that are learned end-to-end. 

\vspace{1mm}
\noindent \textbf{Generalizations.}
PAC, despite being a simple modification to standard convolution,
generalizes several widely used filtering operations, including
\begin{itemize}
    \item \emph{Spatial Convolution} can be seen as a special
    case of PAC with adapting kernel being constant
    $K(\mathbf{f}_i, \mathbf{f}_j) = 1$. This can be achieved
    by using constant adapting features, 
    $\mathbf{f}_i = \mathbf{f}_j, \forall i,j$.
    In brief, standard convolution (Eq.~\ref{eq:conv}) uses 
    fixed, spatially shared filters, while PAC allows the 
    filters to be modified by the adapting kernel $K$ 
    differently across pixel locations.
    \item \emph{Bilateral Filtering}~\cite{tomasi1998bilateral} 
    is a basic image
    processing operation that has found wide-ranging uses~\cite{paris2009bilateral}
    in image processing, computer vision
    and also computer graphics. Standard bilateral
    filtering operation can be seen as a special case
    of PAC, where $\mathbf{W}$ also has a fixed parametric form,
    such as a 2D Gaussian filter,
    $\mathbf{W}\left[\mathbf{p}_i - \mathbf{p}_j\right]=\exp 
    (-\frac{1}{2}(\mathbf{p}_i-\mathbf{p}_j)^\intercal 
    \Sigma^{-1}(\mathbf{p}_i-\mathbf{p}_j))$.
    \item \emph{Pooling} operations can also be modeled
    by PAC. Standard average pooling corresponds to the 
    special case of PAC where 
    $K(\mathbf{f}_i, \mathbf{f}_j)=1,\; \mathbf{W}=\frac{1}{s^2}\cdot 
    \mathbf{1}$. 
    \emph{Detail Preserving 
    Pooling}~\cite{saeedan2018detail,weber2016rapid} is a 
    recently proposed pooling layer that is useful to
    preserve high-frequency details when performing pooling
    in CNNs. PAC can model the detail-preserving 
    pooling operations by incorporating an adapting kernel that 
    emphasizes more distinct pixels in the neighborhood, \eg 
    $K(\mathbf{f}_i, \mathbf{f}_j)=\alpha + 
    \left(|\mathbf{f}_i-\mathbf{f}_j|^2+\epsilon^2\right)^
    \lambda$. 
\end{itemize}

The above generalizations show the generality and the
wide applicability of PAC in different settings and
applications. We experiment using PAC in
three different problem scenarios, 
which will be discussed in later sections. 

Some filtering operations are even more general
than the proposed PAC. Examples include high-dimensional
filtering shown in Eq.~\ref{eq:nd-conv} and others
such as dynamic filter networks (DFN)~\cite{jia2016dynamic} discussed
in Sec.~\ref{sec:related}.
Unlike most of those general filters,
PAC allows efficient learning and reuse of spatially
invariant filters because it is a direct modification of
standard convolution filters.
PAC offers a good trade-off between standard convolution
and DFNs. In DFNs, filters are solely generated by an
auxiliary network and different auxiliary networks or layers
are required to predict kernels for different dynamic-filter
layers. PAC, on the other hand, uses
learned pixel embeddings $\mathbf{f}$ as adapting features, which
can be reused across several different PAC layers in a network.
When related to sparse high-dimensional filtering in
Eq.~\ref{eq:nd-conv}, PAC can be seen as factoring the 
high-dimensional filter into a product of standard spatial
filter $\mathbf{W}$ and the adapting kernel $K$. This allows
efficient implementation of PAC in 2D space alleviating the
need for using hash tables and special lattice structures
in high dimensions. 
PAC can also use learned pixel embeddings $\mathbf{f}$
instead of hand-specified ones in existing learnable high-dimensional
filtering techniques such as  \cite{jampani2016learning}.

\vspace{1mm}
\noindent \textbf{Implementation and variants.}
We implemented PAC as a network layer in PyTorch with GPU 
acceleration\footnote{Code will be available at \url{https://suhangpro.github.io/pac/}}. Our implementation enables back-propagation
through the features $\mathbf{f}$, permitting the
use of learnable deep features as adapting features.
We also implement a PAC variant that does 
pixel-adaptive transposed convolution 
(also called ``deconvolution"). 
We refer to pixel-adaptive convolution shown in 
Eq.~\ref{eq:pac} as PAC and the transposed counterpart as
PAC$^\intercal$. \DS{up to now, we haven't talked about striding for PAC.}
Similar to standard transposed convolution, PAC$^\intercal$
uses fractional striding and results in an upsampled
output.
Our PAC and PAC$^\intercal$ implementations allow
easy and flexible specification of different options that are commonly used in standard convolution:
filter size, number of input and output channels, striding,
padding and dilation factor.


\hs{Comparing runtime with BCL, standard conv, and PAC, similar to~\cite{jampani2016learning} Tab.~3.}

%% file: upsample.tex
\section{Deep Joint Upsampling Networks}
\label{sec:upsampling}
\vspace{-2mm}

Joint upsampling is the task of upsampling a low-resolution
signal with the help of a corresponding high-resolution guidance
image. An example is upsampling a low-resolution depth map given a corresponding high-resolution RGB image as guidance. 
Joint upsampling is useful when some sensors output at a lower resolution than cameras, or can be used to speed up computer vision applications where full-resolution results are expensive to produce.
PAC allows filtering operations to be guided by the adapting features, which can be obtained from a separate guidance image, making it an ideal choice for joint image processing. We investigate the use of PAC for joint upsampling applications. 
In this section, we introduce a network architecture that relies on PAC for deep joint upsampling, and show experimental results on two applications: joint depth upsampling and joint optical flow upsampling.

\subsection{Deep joint upsampling with PAC}
\vspace{-1mm}

\begin{figure}[t!]
\begin{center}
\includegraphics[trim={0.5cm 2cm 0.5cm 3.5cm},clip,width=0.99\columnwidth]{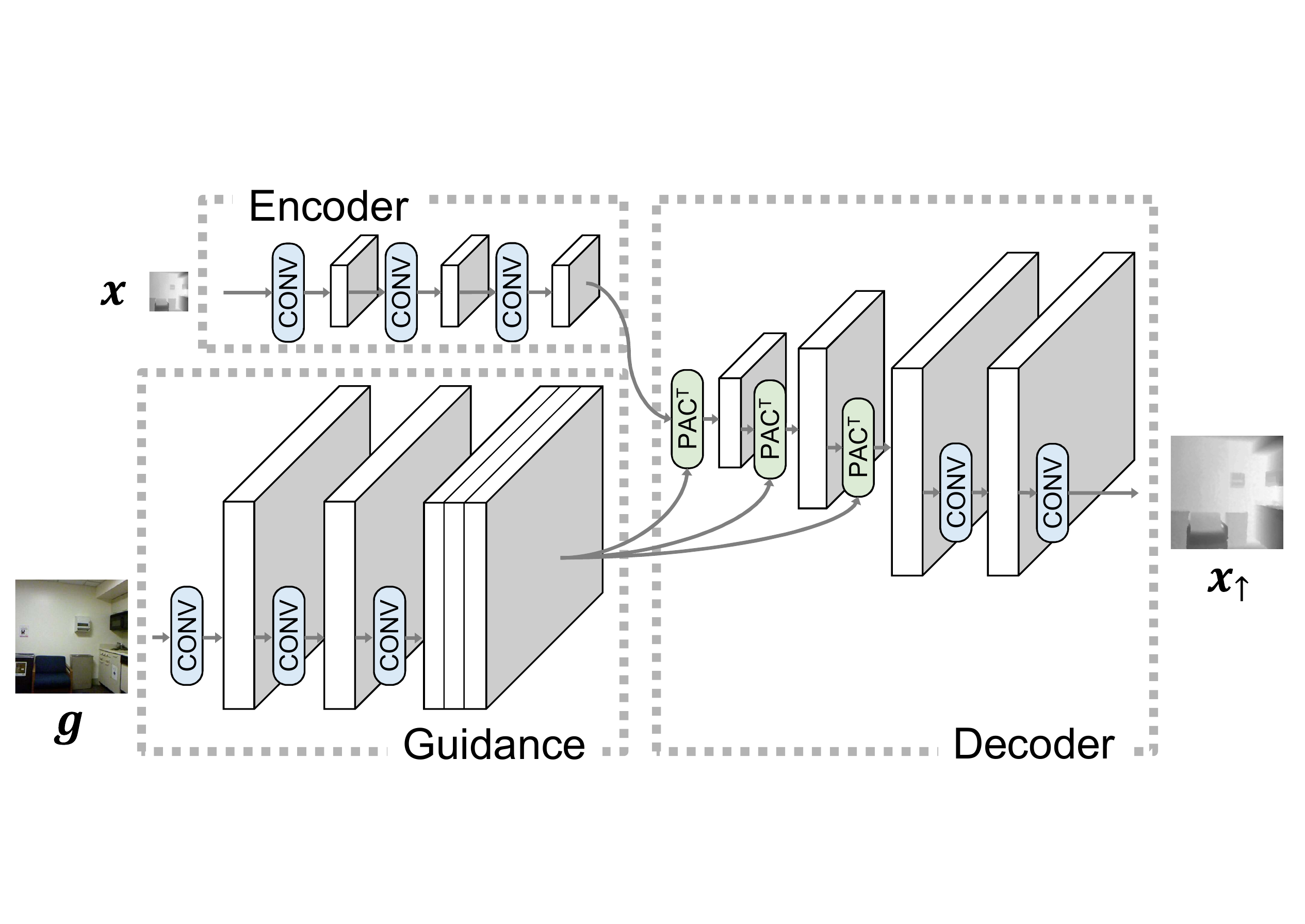} 
  \mycaption{Joint upsampling with PAC}{Network architecture showing encoder, guidance and decoder components. Features from the guidance branch are used to adapt PAC$^\intercal$
  kernels that are applied on the encoder output resulting in
  upsampled signal.}\label{fig:upsample}
\end{center}
\end{figure}

\newcommand{\factorUP}{.14}
\begin{figure*}[t!]
\captionsetup[subfigure]{labelformat=empty}
\begin{centering}
\subfloat{\includegraphics[width=\factorUP\textwidth]{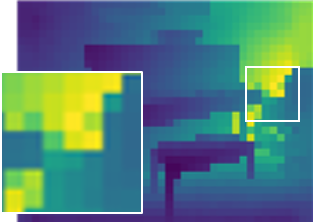}}\quad
\subfloat{\includegraphics[width=\factorUP\textwidth]{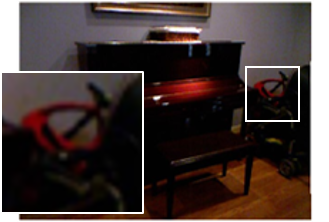}}\quad
\subfloat{\includegraphics[width=\factorUP\textwidth]{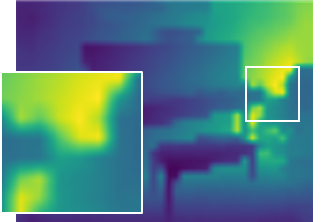}}\quad
\subfloat{\includegraphics[width=\factorUP\textwidth]{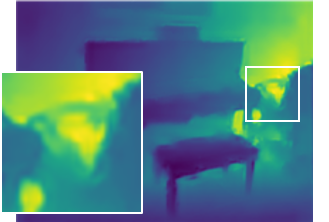}}\quad
\subfloat{\includegraphics[width=\factorUP\textwidth]{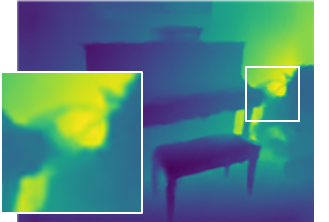}}\quad
\subfloat{\includegraphics[width=\factorUP\textwidth]{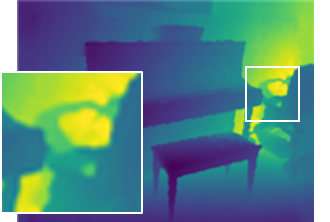}}\\ \vspace{-2mm}
\setcounter{subfigure}{0}
\subfloat[Input]{\includegraphics[width=\factorUP\textwidth]{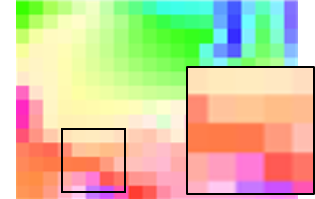}}\quad
\subfloat[Guide]{\includegraphics[width=\factorUP\textwidth]{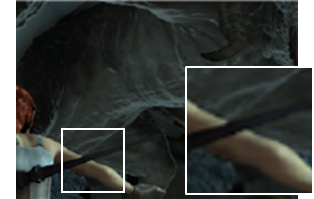}}\quad
\subfloat[Bilinear]{\includegraphics[width=\factorUP\textwidth]{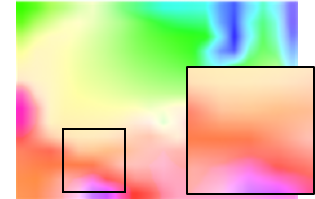}}\quad
\subfloat[DJF]{\includegraphics[width=\factorUP\textwidth]{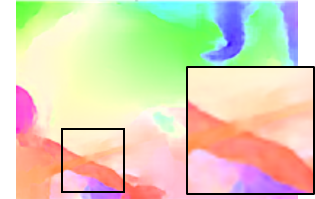}}\quad
\subfloat[Ours]{\includegraphics[width=\factorUP\textwidth]{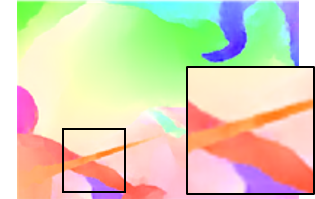}}\quad
\subfloat[GT]{\includegraphics[width=\factorUP\textwidth]{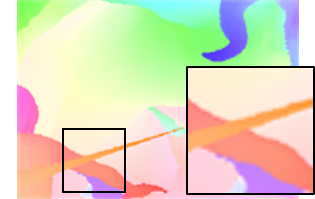}}\\
\mycaption{Deep joint upsampling}{Results of different methods for $16\times$ joint depth upsampling (top row) and $16\times$ joint optical flow upsampling (bottom row). Our method produces results that have more details and are more faithful to the edges in the guidance image.}\label{fig:usampling_results}
\vspace{5mm}
\end{centering}
\end{figure*}

A deep joint upsampling network takes two inputs, a low-resolution signal $\mathbf{x}\in \mathbb{R}^{c \times h/m \times w/m}$ and a high-resolution guidance $\mathbf{g}\in \mathbb{R}^{c_g \times h \times w}$, and outputs upsampled signal $\mathbf{x}_{\uparrow} \in \mathbb{R}^{c \times h\times w}$. Here $m$ is the required upsampling factor. Similar to \cite{li2016deep}, our upsampling network has three components (as illustrated in Fig.~\ref{fig:upsample}): 
\begin{itemize}
    \item \emph{Encoder} branch operates directly on the low-resolution signal with convolution (CONV) layers. 
    \item \emph{Guidance} branch operates solely on the guidance image, and generates adapting features that will be used in all PAC$^\intercal$ layers later in the network. 
    \item \emph{Decoder} branch starts with a sequence of PAC$^\intercal$, which perform transposed pixel-adaptive convolution, each of which upsamples the feature maps by a factor of 2.
    PAC$^\intercal$ layers are followed by two CONV layers to generate the 
    final upsampled output.
\end{itemize}
Each of the CONV and PAC$^\intercal$ layers, except the final one, is followed by a rectified linear unit (ReLU).

\begin{table}[b!]
    \vspace{-2mm}
    \centering
     \mycaption{Joint depth upsampling}{Results (in RMSE) show that our upsampling network consistently outperforms other techniques for different upsampling factors.}
    \small
    \begin{tabular}{l c c c }
        \toprule
        Method & 4$\times$ & 8$\times$ & 16$\times$  \\
        \midrule
        Bicubic & 8.16 & 14.22 & 22.32  \\
        MRF	& 7.84 & 13.98 & 22.20 \\ 
        GF~\cite{he2013guided}	& 7.32 & 13.62 & 22.03 \\ 
        JBU~\cite{kopf2007joint} & 4.07 & 8.29 & 13.35 \\ 
        Ham \etal~\cite{ham2015robust} & 5.27 & 12.31 & 19.24 \\ 
        DMSG~\cite{hui2016depth} & 3.78 & 6.37 & 11.16 \\ 
        FBS~\cite{barron2016fast} & 4.29 & 8.94 & 14.59 \\ 
        DJF~\cite{li2016deep} & 3.54 & 6.20 & 10.21  \\ 
        DJF+~\cite{li2018joint} & 3.38 & 5.86 & 10.11  \\
        DJF (Our impl.) & 2.64 & 5.15 & 9.39 \\ 
        \midrule
        Ours-lite & 2.55 & 4.82 & 8.52 \\ 
        Ours & \textbf{2.39} & \textbf{4.59} & \textbf{8.09} \\ 
        \bottomrule
    \end{tabular}
    \label{tab:upsample}
\end{table}

\subsection{Joint depth upsampling}
\label{sec:upsample_depth}
\vspace{-1mm}

Here, the task is to upsample a low-resolution depth by using a
high-resolution RGB image as guidance.
We experiment with the NYU Depth V2 dataset~\cite{silberman2012indoor}, which has 
1449 RGB-depth pairs. Following~\cite{li2016deep}, we use the first 1000 samples for training and the rest for testing. The low-resolution depth maps are obtained from the ground-truth depth maps using nearest-neighbor downsampling. 
Tab.~\ref{tab:upsample} shows root
mean square error (RMSE) of different techniques and for different upsampling factors $m$ (4$\times$, 8$\times$, 16$\times$).
Results indicate that our network outperforms others in comparison and obtains state-of-the-art performance.
Sample visual results are shown in Fig.~\ref{fig:usampling_results}. 

We train our network with the Adam optimizer using a learning rate schedule of [$10^{-4} \times$ 3.5k, $10^{-5} \times$ 1.5k, $10^{-6}  \times$ 0.5k] and with mini-batches of 256$\times$256 crops. We found this training setup to be superior to the one recommended in DJF~\cite{li2016deep}, and also compare with our own implementation of it under such a setting (``DJF (Our impl.)" in Tab.~\ref{tab:upsample}). 
We keep the network architecture similar to that of previous state-of-the-art technique, DJF~\cite{li2016deep}. In DJF, features from the guidance
branch  are simply concatenated with encoder outputs for upsampling, whereas we use guidance features to adapt PAC$^\intercal$ kernels.
Although with similar number of layers, our network has more parameters compared with DJF (see \supp{} for details). We also trained a lighter version of our network (``Ours-lite") that matches the number of parameters of DJF, and still observe better performance showing the importance of PAC$^\intercal$ for upsampling.

\subsection{Joint optical flow upsampling}
\vspace{-1mm}

We also evaluate our joint upsampling network for upsampling low-resolution optical flow using the original RGB image as 
guidance. 
Estimating optical flow is a challenging task, and even recent state-of-the-art approaches~\cite{sun2018pwc} resort to
simple bilinear upsampling to predict optical flow at the full resolution. 
Optical flow is smoothly varying within motion boundaries, where accompanying RGB images can offer strong clues, making joint upsampling an appealing solution. 
We use the same network architecture as in the depth upsampling experiments, with the only difference being that instead of
single-channel depth, input and output are two-channel flow with $u,v$ components. We experiment with the Sintel dataset~\cite{butler2012sintel} (clean pass). 
The same training protocol in Sec.~\ref{sec:upsample_depth} is used, and the low-resolution optical flow is obtained from bilinear downsampling of the ground-truth. We compare with baselines of bilinear interpolation and 
DJF~\cite{li2016deep}, and observe consistent advantage (Tab.~\ref{tab:upsample_flow}). Fig.~\ref{fig:usampling_results} shows a
sample visual result indicating that our network is capable of restoring fine-structured details and also produces smoother predictions in areas with uniform motion. 

\begin{table}[hb!]
    \vspace{4mm}
    \mycaption{Joint optical flow upsampling}{End-Point-Error (EPE) showing the improved performance compared with DJF~\cite{li2016deep}.}
    \centering
    \small
    \begin{tabular}{l c c c }
    \toprule
          & 4$\times $ & 8$\times$ & 16$\times$ \\
    \midrule
    Bilinear & 0.465 & 0.901 & 1.628 \\ 
    DJF~\cite{li2016deep} & 0.176	& 0.438	& 1.043 \\ 
    \midrule
    Ours & \textbf{0.105} & \textbf{0.256} & \textbf{0.592} \\ 
    \bottomrule
    \end{tabular}
    \label{tab:upsample_flow}
\end{table}

%% file: crf.tex
\section{Conditional Random Fields}
\label{sec:crf}
\vspace{-2mm}

Early adoptions of CRFs in computer vision tasks were limited to region-based approaches and short-range structures~\cite{shotton2006textonboost} for efficiency reasons. Fully-Connected CRF (Full-CRF) \cite{krahenbuhl2011efficient} was proposed to offer the benefits of dense pairwise connections among pixels, which resorts to approximate high-dimensional filtering~\cite{adams2010fast} for efficient inference.
Consider a semantic labeling problem, where each 
pixel $i$ in an image $I$ can take one of the semantic labels $l_i \in \{1,...,\mathcal{L} \}$. Full-CRF has unary potentials
usually defined by a classifier such as CNN: 
$\psi_u(l_i) \in \mathbb{R}^{\mathcal{L}}$. And, the
pairwise potentials are defined for every pair of pixel locations $(i, j)$:
$ \psi_p(l_i, l_j|I) = \mu(l_i, l_j)K(\mathbf{f}_i, \mathbf{f}_j)$,
where $K$ is a kernel function and $\mu$ is a compatibility function. A common choice for $\mu$ is the Potts model: $\mu(l_i, l_j) = [l_i\neq l_j]$. \cite{krahenbuhl2011efficient} utilizes two Gaussian kernels with hand-crafted features as the kernel function: 
\begin{align}
    K(\mathbf{f}_i, \mathbf{f}_j) = &w_1 \exp \left\{-\frac{\|\mathbf{p}_i - \mathbf{p}_j\|^2}{2\theta_\alpha^2}-\frac{\|I_i-I_j\|^2}{2\theta_\beta^2}\right\} \nonumber \\ 
    &+ w_2 \exp \left\{ -\frac{\|\mathbf{p}_i - \mathbf{p}_j\|^2}{2\theta_\gamma^2} \right\} \label{eq:k_fullcrf}
\end{align}
where $w_1, w_2, \theta_\alpha, \theta_\beta, \theta_\gamma$ are model parameters, and are typically found by a grid-search. 
Then, inference in Full-CRF amounts to maximizing the 
following Gibbs distribution: $P(\mathbf{l}|I) \!=\! \exp(-\sum_i \psi_u(l_i) \!-\! \sum_{i < j} \psi_p(l_i, l_j))$, 
$\mathbf{l} = (l_1,l_2,...,l_n)$. Exact inference of Full-CRF is hard, and~\cite{krahenbuhl2011efficient} relies on mean-field approximation which is optimizing for an approximate
distribution $Q(\mathbf{l})=\prod_i Q_i(l_i)$ by
minimizing the KL-divergence between $P(\mathbf{l}|I)$ and the mean-field approximation $Q(\mathbf{l})$. This leads to the following mean-field (MF) inference step that updates marginal distributions $Q_i$ iteratively for $t=0,1,...$ : 
\begin{align}
    Q_i^{(t+1)}(l) \leftarrow &\frac{1}{Z_i} \exp \biggr\{ -\psi_u (l) \nonumber\\ &- \sum_{l' \in \mathcal{L}} \mu(l, l')\sum_{j\neq i}K(\mathbf{f}_i, \mathbf{f}_j) Q_j^{(t)}(l') \biggr\}  \label{eq:update_fullcrf}
\end{align}


The main computation in each MF iteration, $\sum_{j\neq i}K(\mathbf{f}_i, \mathbf{f}_j) Q_j^{(t)}$, can be viewed as high-dimensional Gaussian filtering. Previous work~\cite{krahenbuhl2011efficient, krahenbuhl2013parameter} relies on permutohedral lattice convolution~\cite{adams2010fast} to achieve efficient implementation. 

\begin{figure}[!ht]
    \vspace{-1mm}
    \centering
    \includegraphics[trim={1.2cm 4.9cm 6.2cm 5cm},clip,width=0.95\columnwidth]{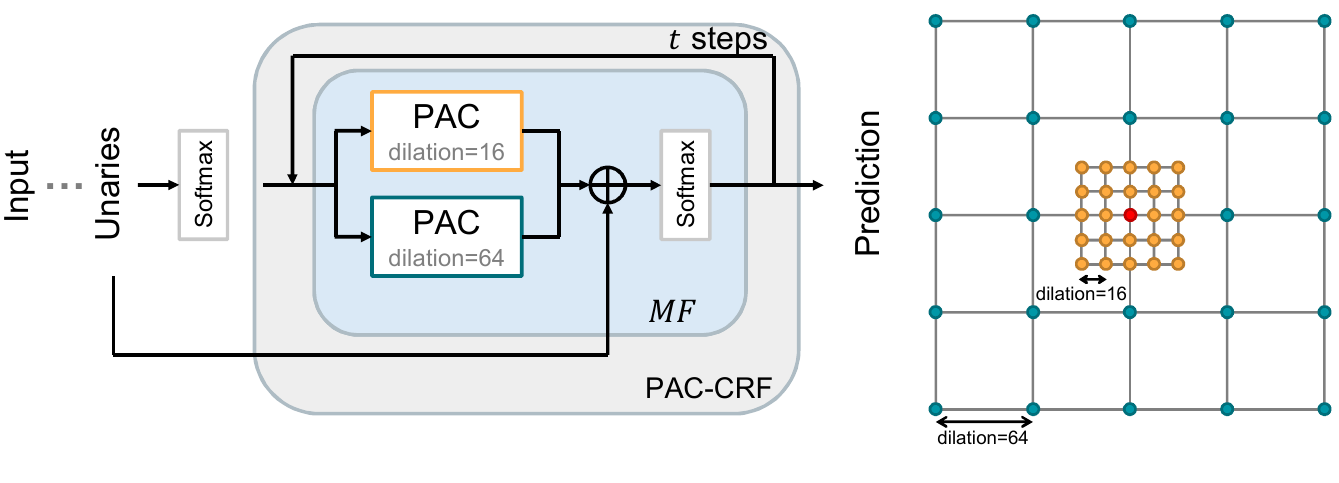}
    \mycaption{PAC-CRF}{Illustration of inputs, outputs and the operations in each mean-field (MF) step of PAC-CRF inference. Also shown is the coverage of two $5\times5$ PAC filters, with dilation factors 16 and 64 respectively.}
    \vspace{0.5cm}
    \label{fig:crf}
\end{figure}

\subsection{Efficient, learnable CRF with PAC}
\label{sec:convcrf}
\vspace{-1mm}

Existing work~\cite{zheng2015conditional,jampani2016learning} back-propagates through the above MF steps to combine CRF
inference with CNNs resulting in end-to-end training of CNN-CRF
models.
While there exists optimized CPU implementations, permutohedral lattice convolution cannot easily utilize GPUs because it ``does not follow the SIMD paradigm of efficient GPU computation" \cite{Teichmann2018convolutional}. Another drawback of relying on permutohedral lattice convolution is the approximation error incurred during both inference and gradient computation. 

We propose PAC-CRF, which alleviates these computation issues by relying on PAC for efficient inference, and is easy to integrate with existing CNN backbones. PAC-CRF also has additional learning capacity, which leads to better performance compared with Full-CRF in our experiments. 

\vspace{1mm}
\noindent \textbf{PAC-CRF.} 
In PAC-CRF, we define pairwise connections over fixed windows
$\Omega^k$ around each pixel instead of dense connections:
$\sum_k\sum_i\sum_{j\in\Omega^k(i)} \psi^k_p (l_i, l_j | I)$,
where the $k$-th pairwise potential is defined as
\begin{align}
    \psi_p^k(l_i, l_j|I) = K^k(\mathbf{f}_i,\mathbf{f}_j) \mathbf{W}_{l_j l_i}^k[\mathbf{p}_j-\mathbf{p}_i]  \label{eq:pairwise_paccrf}
\end{align}
Here $\Omega^k(\cdot)$ specifies the pairwise connection pattern of the $k$-th pairwise potential originated from each pixel, and $K^k$ is a fixed Gaussian kernel. 
Intuitively, this formulation allows the label compatibility transform $\mu$ in Full-CRF to be modeled by $\mathbf{W}$, and to vary across different spatial offsets. 
Similar derivation as in Full-CRF yields the following 
iterative MF update rule (see \supp{} for more details): 
\begin{align}
    &Q_i^{(t+1)}(l) \leftarrow \frac{1}{Z_i} \exp \biggr\{ -\psi_u (l) - \nonumber\\ &\sum_k\underbrace{\sum_{l' \in \mathcal{L}}\sum_{j\in \Omega^k(i)}K^k(\mathbf{f}_i, \mathbf{f}_j) \mathbf{W}^k_{l'l}[\mathbf{p}_j-\mathbf{p}_i] Q_j^{(t)}(l')}_\text{PAC} \biggr\} \label{eq:update_paccrf}
\end{align}

MF update now consists of PAC instead of sparse high-dimensional
filtering as in Full-CRF (Eq.~\ref{eq:update_fullcrf}). As outlined
in Sec.~\ref{sec:related}, there are several advantages
of PAC over high-dimensional filtering. With PAC-CRF, we
can freely parameterize and learn the pairwise potentials in Eq.~\ref{eq:pairwise_paccrf} that also use a richer form of
compatibility transform~$\mathbf{W}$. PAC-CRF can also make use of
learnable features $\mathbf{f}$ for pairwise potentials
instead of pre-defined ones in Full-CRF.
Fig.~\ref{fig:crf} (left) illustrates the computation steps in each
MF step with two pairwise PAC kernels.

\vspace{1mm}
\noindent \textbf{Long-range connections with dilated PAC.}
The major source of heavy computation in Full-CRF is the dense pairwise pixel connections. In PAC-CRF, the pairwise connections are defined by the local convolution windows $\Omega^k$.
To have long-range pairwise connections while keeping the
number of PAC parameters managable, we make use of dilated filters~\cite{chen2018deeplab,yu2015multi}. 
Even with a relatively small kernel size ($5\times 5$), with a large dilation, \eg $64$, the CRF can effectively reach a neighborhood of $257\times 257$. A concurrent work~\cite{Teichmann2018convolutional} also propose a 
convolutional version of CRF (Conv-CRF) 
to reduce the number of connections in Full-CRF. However,  \cite{Teichmann2018convolutional} uses connections only
within small local windows. We argue that long-range connections can provide valuable information, and our CRF formulation uses a wider range of connections while still being efficient. Our formulation allows using multiple PAC filters in parallel, each with different dilation factors. In Fig.~\ref{fig:crf} (right), we show an illustration of the coverage of two $5\times5$ PAC filters, with dilation factors 16 and 64 respectively. This allows PAC-CRF to achieve a good trade-off between computational efficiency and
long-range pairwise connectivity.

\subsection{Semantic segmentation with PAC-CRF}
\vspace{-2mm}

The task of semantic segmentation is to assign a semantic label
to each pixel in an image. Full-CRF is proven to be a valuable
post-processing tool that can considerably
improve CNN segmentation performance~\cite{chen2018deeplab,zheng2015conditional,jampani2016learning}.
Here, we experiment with PAC-CRF on top of the FCN semantic segmentation network~\cite{long2015fully}. We
choose FCN for simplicity and ease of comparisons, as 
FCN only uses standard convolution layers and does not have 
many bells and whistles.

In the experiments, we use scaled RGB color, $[\frac{R}{\sigma_R},\frac{G}{\sigma_G}, \frac{B}{\sigma_B}]^\intercal$, as the guiding features for the PAC layers in PAC-CRF . The scaling vector $[\sigma_R, \sigma_G, \sigma_B]^\intercal$ is learned jointly with the PAC weights $\mathbf{W}$. We try two internal configurations of PAC-CRF: 
a single 5$\times$5 PAC kernel with dilation of 32, and two parallel 5$\times$5 PAC kernels with dilation factors of 16 and 64.
5 MF steps are used for a good balance between speed and accuracy (more details in \supp). 
We first freeze the backbone FCN network and train only the PAC-CRF part for 40 epochs, and then train the whole network for another 40 epochs with reduced learning rates. 

\vspace{1mm}
\noindent \textbf{Dataset.} We follow the training and validation
settings of FCN~\cite{long2015fully} 
which is trained on PascalVOC
images and validated on a reduced validation set of 736 images.
We also submit our final trained models to the official evaluation server to get test scores on 1456 test images.

\vspace{1mm}
\noindent \textbf{Baselines.} We compare PAC-CRF with three baselines: Full-CRF~\cite{krahenbuhl2011efficient},  BCL-CRF~\cite{jampani2016learning}, and Conv-CRF~\cite{Teichmann2018convolutional}. For Full-CRF, we use the publicly available C++ code, and find the optimal CRF parameters through grid search.
For BCL-CRF, we use $1$-neighborhood filters to keep the runtime manageable and use other settings as suggested by the authors. 
For Conv-CRF, the same training procedure is used as in PAC-CRF. We use the more powerful variant of Conv-CRF with learnable compatibility transform (referred to as ``Conv+C'' in~\cite{Teichmann2018convolutional}), and we learn the RGB scales for Conv-CRF in the same way as for PAC-CRF.
We follow the suggested default settings for Conv-CRF and use a filter size of 11$\times$11 and a blurring factor of 4. Note that like Full-CRF (Eq.~\ref{eq:k_fullcrf}), the other baselines also use two pairwise kernels. 

\begin{table}[t!]
    \centering
    \vspace{2mm}
    \mycaption{Semantic segmentation with PAC-CRF}{Validation
    and test mIoU scores along with the runtimes of different techniques. PAC-CRF results in better improvements than
    Full-CRF~\cite{krahenbuhl2011efficient} 
    while being faster. PAC-CRF also outperforms
    Conv-CRF~\cite{Teichmann2018convolutional} and BCL~\cite{jampani2016learning}. Runtimes are averaged over all validation images.}
    \small
    \begin{tabular}{l l c}
        \toprule
        Method & mIoU (val / test) & CRF Runtime  \\
        \midrule
         Unaries only (FCN) & 65.51 / 67.20 & - \\
        \midrule
        
        Full-CRF~\cite{krahenbuhl2011efficient} & +2.11 / +2.45 & 629 ms \\
        BCL-CRF~\cite{jampani2016learning} & +2.28 / +2.33 & 2.6 s \\
        Conv-CRF~\cite{Teichmann2018convolutional} & +2.13 / +1.57 & 38 ms\\
        \midrule
        PAC-CRF, 32 & +3.01 / +2.21 & 39 ms \\ 
        PAC-CRF, 16-64 & \textbf{+3.39} / \textbf{+2.62} & 78 ms \\ 
        \bottomrule
    \end{tabular}

    \vspace{-2mm}
    \label{tab:crf}
\end{table}

\vspace{1mm}
\noindent \textbf{Results.}
Tab.~\ref{tab:crf} reports validation and test mean
Intersection over Union (mIoU) scores along with average runtimes of different techniques.
Our two-filter variant (``PAC-CRF, 16-64'') achieves better mIoU compared with all baselines, and also compares favorably in terms of runtime. The one-filter variant (``PAC-CRF, 32'') performs slightly worse than Full-CRF and BCL-CRF, but has even larger speed advantage, offering a strong option where efficiency is needed. Sample visual results are shown in Fig.~\ref{fig:crf_result}. While being quantitatively better and retaining more visual details overall, PAC-CRF produces some amount of noise around boundaries. This is likely due to a known ``gridding" effect of dilation~\cite{yu2017dilated}, which we hope to mitigate in future work.

\newcommand{\factor}{0.0865}
\newcommand{\capex}{1.1}
\begin{figure*}
\captionsetup[subfigure]{labelformat=empty}
\begin{centering}
\subfloat{\includegraphics[width=\factor\textwidth]{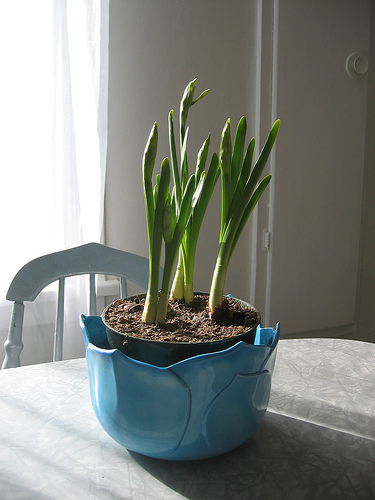}}~
\subfloat{\makebox[\capex \width]{\includegraphics[width=\factor\textwidth]{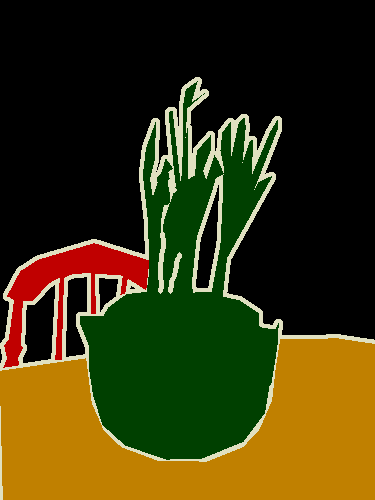}}}~
\subfloat{\includegraphics[width=\factor\textwidth]{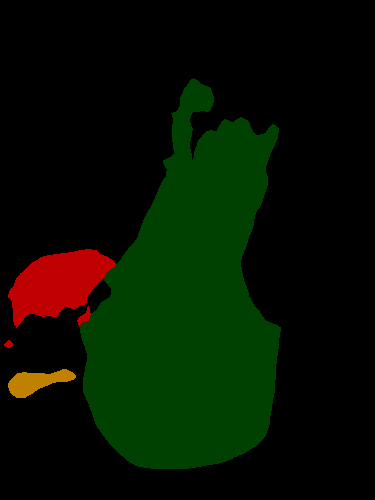}}~
\subfloat{\makebox[\capex \width]{\includegraphics[width=\factor\textwidth]{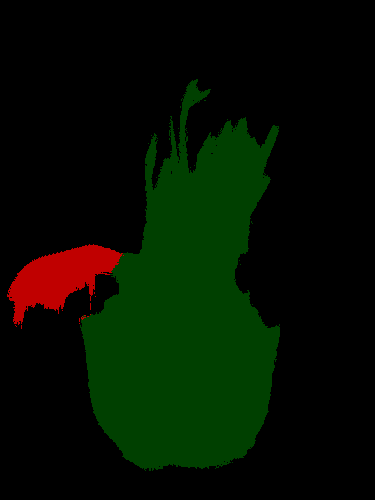}}}~
\subfloat{\includegraphics[width=\factor\textwidth]{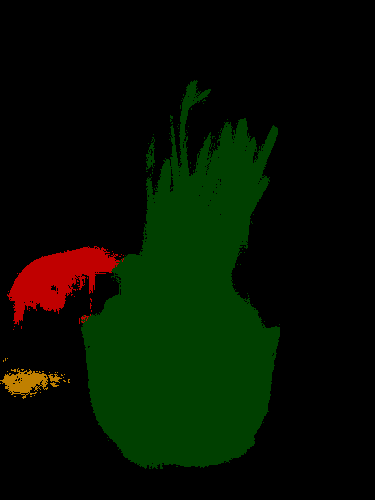}}~
\subfloat{\makebox[\capex \width]{\includegraphics[width=\factor\textwidth]{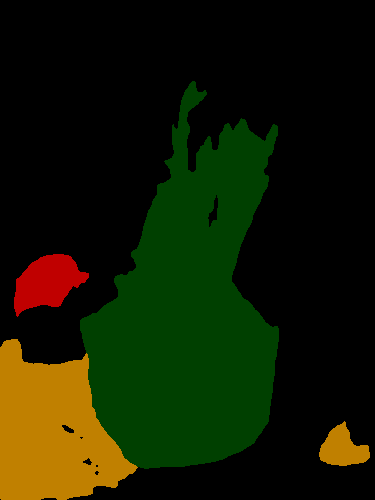}}}~
\subfloat{\includegraphics[width=\factor\textwidth]{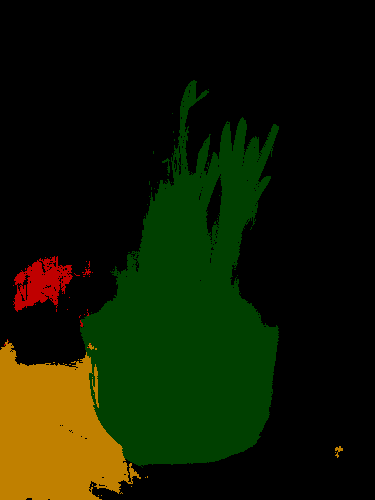}}~
\subfloat{\makebox[\capex \width]{\includegraphics[width=\factor\textwidth]{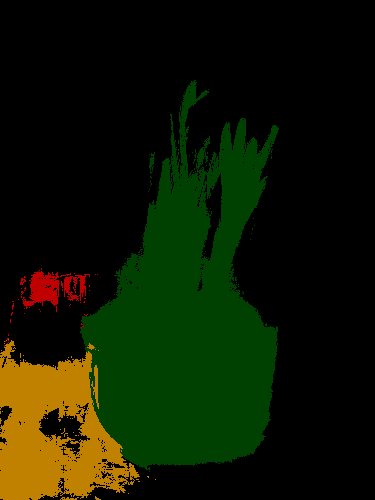}}}~
\subfloat{\includegraphics[width=\factor\textwidth]{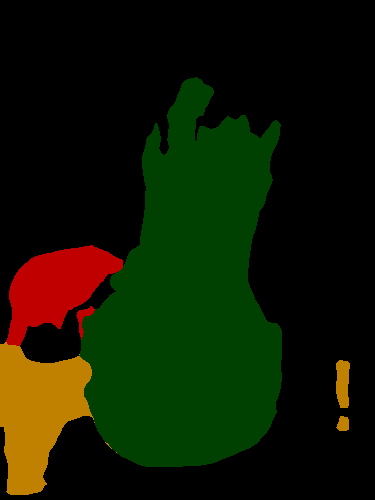}}~
\subfloat{\makebox[\capex \width]{\includegraphics[width=\factor\textwidth]{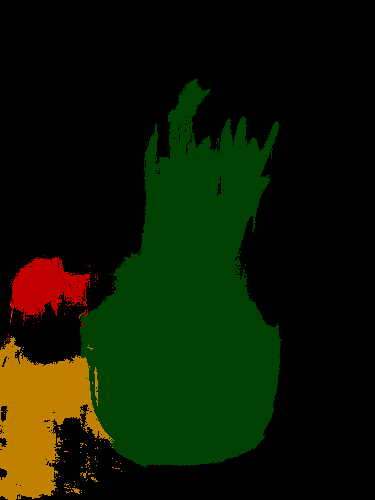}}}~\\
\vspace{-2mm}
\subfloat{\includegraphics[width=\factor\textwidth]{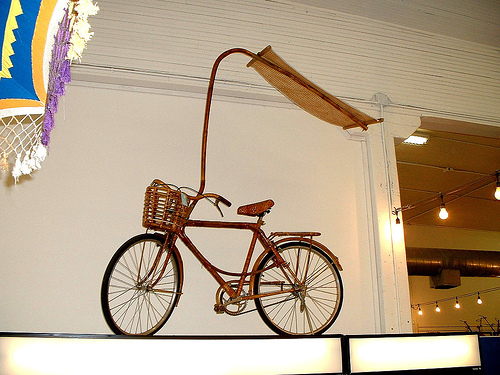}}~
\subfloat{\makebox[\capex \width]{\includegraphics[width=\factor\textwidth]{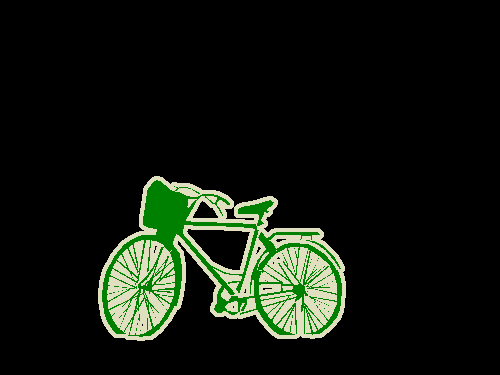}}}~
\subfloat{\includegraphics[width=\factor\textwidth]{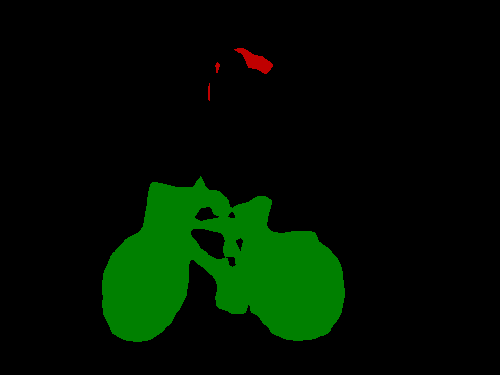}}~
\subfloat{\makebox[\capex \width]{\includegraphics[width=\factor\textwidth]{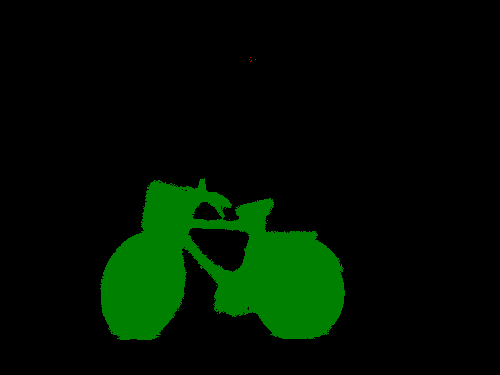}}}~
\subfloat{\includegraphics[width=\factor\textwidth]{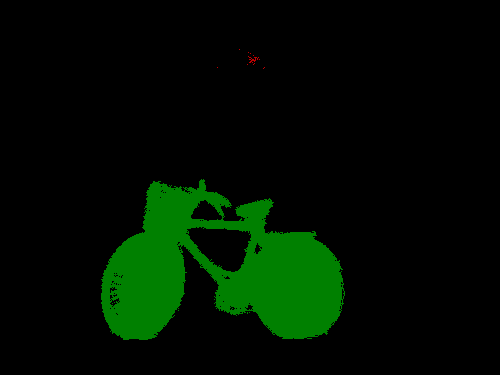}}~
\subfloat{\makebox[\capex \width]{\includegraphics[width=\factor\textwidth]{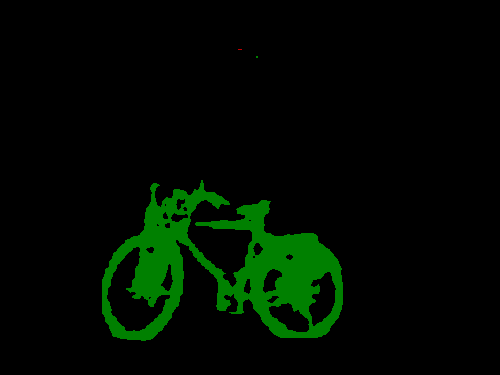}}}~
\subfloat{\includegraphics[width=\factor\textwidth]{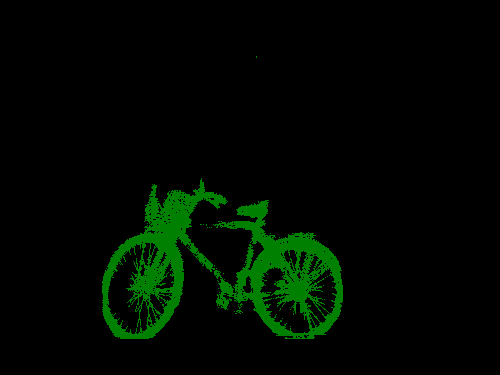}}~
\subfloat{\makebox[\capex \width]{\includegraphics[width=\factor\textwidth]{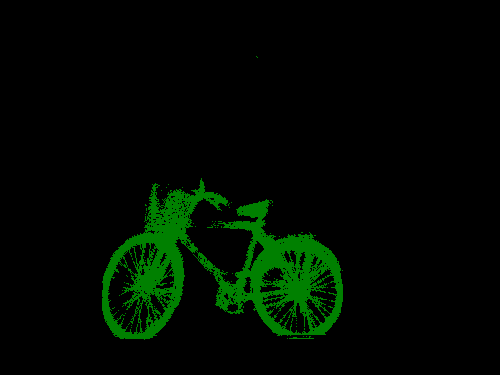}}}~
\subfloat{\includegraphics[width=\factor\textwidth]{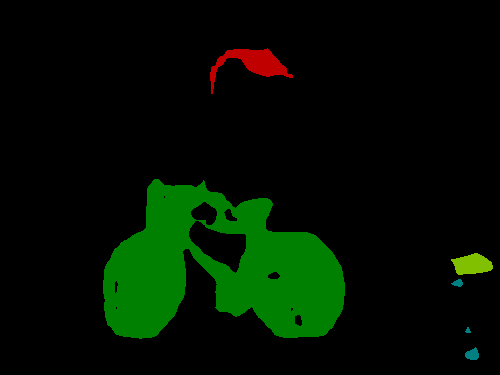}}~
\subfloat{\makebox[\capex \width]{\includegraphics[width=\factor\textwidth]{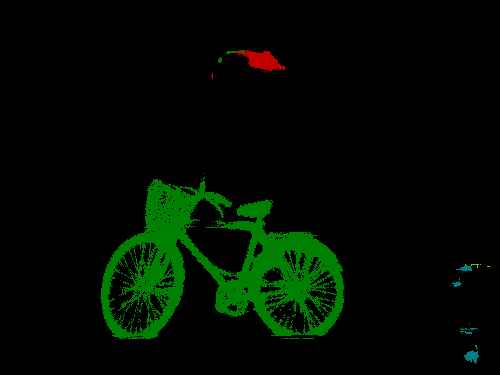}}}~\\
\vspace{-2mm}
\setcounter{subfigure}{0}
\subfloat[\scriptsize Input]{\includegraphics[width=\factor\textwidth]{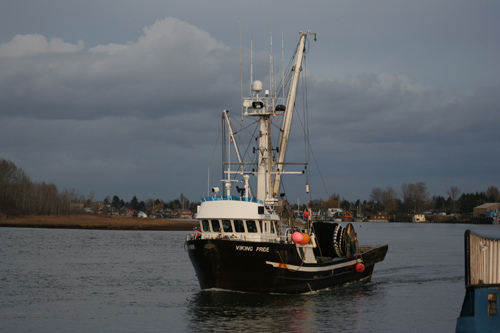}}~
\subfloat[\scriptsize GT]{\makebox[\capex \width]{\includegraphics[width=\factor\textwidth]{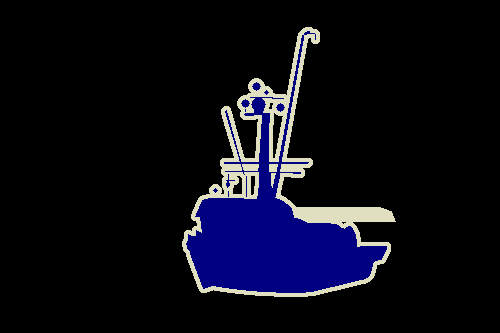}}}~
\subfloat[\scriptsize Unary]{\includegraphics[width=\factor\textwidth]{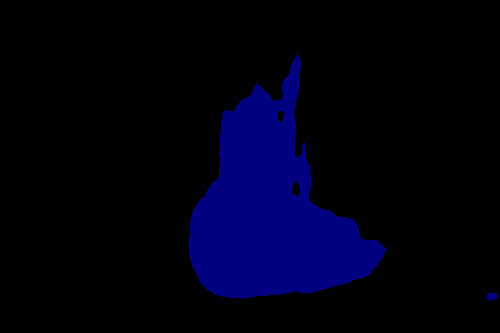}}~
\subfloat[\scriptsize Full-CRF]{\makebox[\capex \width]{\includegraphics[width=\factor\textwidth]{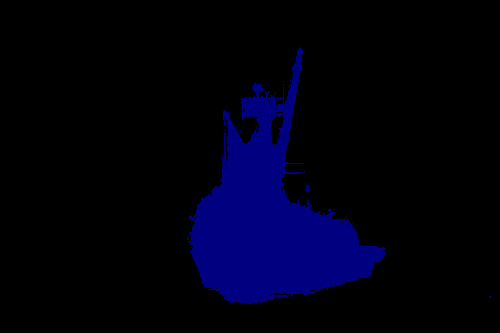}}}~
\subfloat[\scriptsize BCL-CRF]{\includegraphics[width=\factor\textwidth]{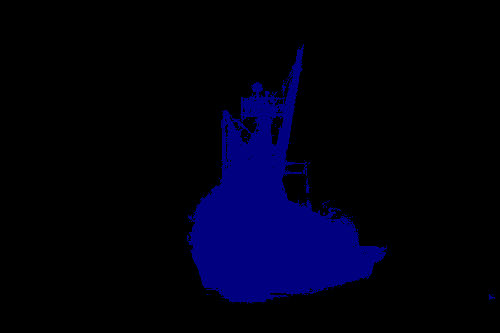}}~
\subfloat[\scriptsize Conv-CRF]{\makebox[\capex \width]{\includegraphics[width=\factor\textwidth]{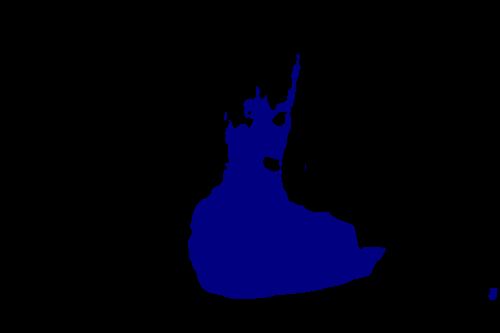}}}~
\subfloat[\scriptsize PAC-CRF,32]{\includegraphics[width=\factor\textwidth]{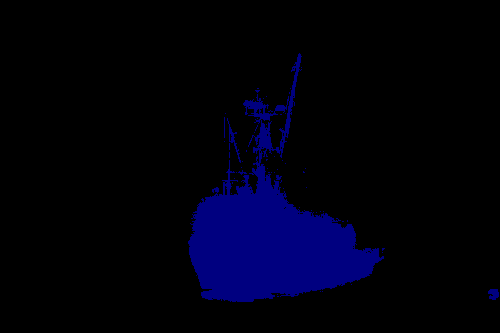}}~
\subfloat[\scriptsize PAC-CRF,16-64]{\makebox[\capex \width]{\includegraphics[width=\factor\textwidth]{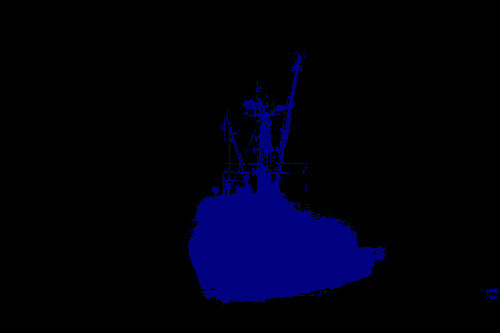}}}~
\subfloat[\scriptsize PAC-FCN]{\includegraphics[width=\factor\textwidth]{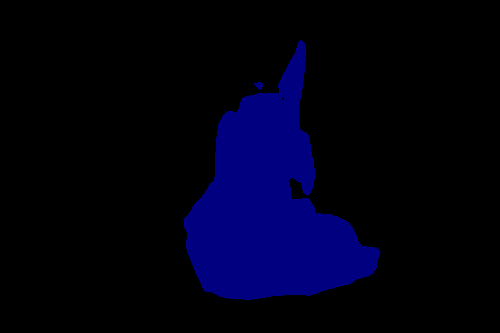}}~
\subfloat[\scriptsize PAC-FCN-CRF]{\makebox[\capex \width]{\includegraphics[width=\factor\textwidth]{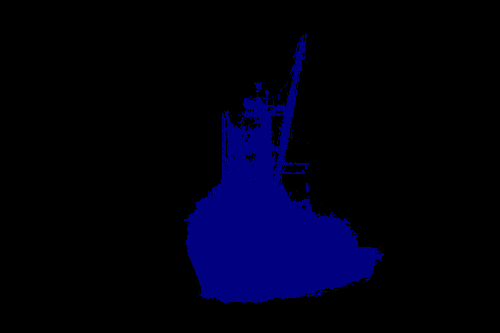}}}~\\
\mycaption{Semantic segmentation with PAC-CRF and PAC-FCN}{We show three examples from the validation set. Compared to Full-CRF~\cite{krahenbuhl2011efficient}, BCL-CRF~\cite{jampani2016learning}, and Conv-CRF~\cite{Teichmann2018convolutional}, PAC-CRF can recover finer details faithful to the boundaries in the RGB inputs. }\label{fig:crf_result}
\vspace{4mm}
\end{centering}
\end{figure*}

%% file: hotswap.tex
\section{Layer hot-swapping with PAC}
\label{sec:hotswap}
\vspace{-2mm}

\begin{figure}[b!]
\vspace{-2mm}
\begin{center}
  \centerline{\includegraphics[trim={1.7cm 1.8cm 1.4cm 1.75cm},clip,width=0.95\columnwidth]{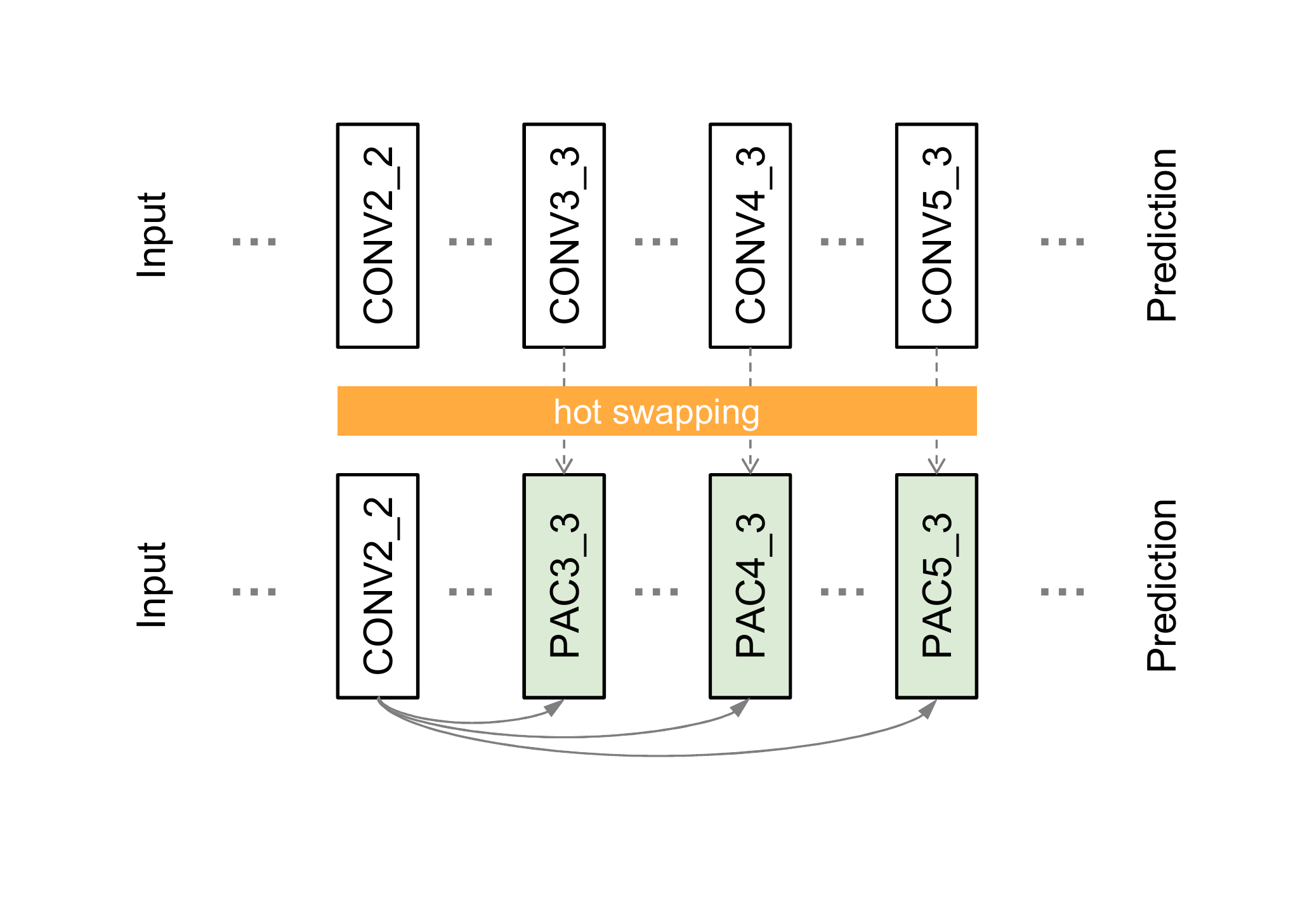}} \vspace{-0.0cm} 
  \mycaption{Layer hot-swapping with PAC}{A few layers of a network before (top) and after (bottom) hot-swapping. Three CONV layers are replaced with PAC layers, with adapting features coming from an earlier convolution layer. All the original network weights are retained after the modification.}
  \label{fig:hotswap}
\end{center}
\end{figure}

So far, we design specific architectures around PAC for different use cases. In this section, we offer a strategy to use PAC for simply upgrading existing CNNs with minimal modifications through what we call layer \emph{hot-swapping}. 

\vspace{1mm}
\noindent \textbf{Layer hot-swapping.} Network fine-tuning has become a common practice when training networks on new data or with additional layers. Typically, in fine-tuning, newly added layers are initialized randomly. Since PAC generalizes standard convolution layers, it can directly replace convolution layers in existing networks while retaining the pre-trained weights. We refer to this modification of existing pre-trained networks as layer \emph{hot-swapping}. 

We continue to use semantic segmentation as an example, and demonstrate how layer hot-swapping can be a simple yet effective modification to existing CNNs. Fig.~\ref{fig:hotswap} illustrates a FCN~\cite{long2015fully} before and after the hot-swapping modifications.~We swap out the last CONV layer of the last three convolution groups, CONV3\_3, CONV4\_3, CONV5\_3, with PAC layers with the same configuration (filter size, input and output channels, \etc), and use the output of CONV2\_2 as the guiding feature for the PAC layers. By this example, we also demonstrate that one could use
earlier layer features (CONV2\_2 here) as adapting features for PAC. Using this strategy, the network parameters do not increase
when replacing CONV layers with PAC layers.
All the layer weights are initialized with trained FCN parameters. To ensure a better starting condition for further training, we scale the guiding features by a small constant (0.0001) so that the PAC layers initially behave very closely to their original CONV counterparts.
We use 8825 images for training, including the Pascal VOC 2011 training images and the additional training samples from~\cite{hariharan2011semantic}. Validation and testing are performed in the same fashion as in Sec.~\ref{sec:crf}. 

Results are reported in Tab.~\ref{tab:hotswap}. We show that our simple modification (PAC-FCN) provides about $2$ mIoU improvement on test ($67.20\rightarrow69.18$) for the semantic segmentation task, while incurring virtually no runtime penalty at inference time. Note that PAC-FCN has the same number of parameters as the original FCN model. The improvement brought
by PAC-FCN is also complementary to any additional CRF post-processing that can still be applied. After combined with a PAC-CRF (the 16-64 variant) and trained jointly, we observe another $2$ mIoU improvement. \OG{Typo, something is missing: ``After SOMETHING IS combined''} Sample visual results are shown
in Fig.~\ref{fig:crf_result}.

\begin{table}[h!]
    \centering
    \mycaption{FCN hot-swapping CONV with PAC}{Validation
    and test mIoU scores along with runtimes of different techniques. Our simple hot-swapping strategy provides $2$ IoU gain on test. Combining with PAC-CRF offers additional improvements.}
    \small
    \begin{tabular}{l c c r}
    \toprule
        Method & PAC-CRF & mIoU (val / test) & Runtime\\
        \midrule
        FCN-8s & - & 65.51 / 67.20 & 39 ms \\
        FCN-8s & 16-64 & 68.90 /  69.82 & 117 ms \\
        \midrule
        PAC-FCN & - &  67.44 / 69.18 & 41 ms \\
        PAC-FCN & 16-64 & \textbf{69.87} /  \textbf{71.34} & 118 ms \\
    \bottomrule
    \end{tabular}
    \label{tab:hotswap}
\end{table}

%% file: conclusion.tex
\section{Conclusion}
\label{sec:conclusion}
\vspace{-2mm}

In this work we propose PAC, a new type of filtering operation that can effectively learn to leverage guidance information. 
We show that PAC generalizes several popular filtering operations and demonstrate its applicability on different uses ranging from joint upsampling, semantic segmentation networks, to efficient CRF inference. PAC generalizes standard spatial convolution, and can be used to directly replace standard convolution layers in pre-trained networks for performance gain with minimal computation overhead. 

%% file: acknowledgements.tex
\paragraph{Acknowledgements} 

Hang Su and Erik Learned-Miller acknowledge support from AFRL and DARPA (\#FA8750-18-2-0126)\footnote{The U.S.\ Gov.\ is authorized to reproduce and distribute reprints for Gov.\ purposes notwithstanding any copyright notation thereon. The views and conclusions contained herein are those of the authors and should not be interpreted as necessarily representing the official policies or endorsements, either expressed or implied, of the AFRL and DARPA or the U.S.\ Gov.} and the MassTech Collaborative grant for funding the UMass GPU cluster.

%% file: supp.tex
\section{Deep Joint Upsampling with PAC}
\label{sec:supp_upsample}
\paragraph{Network architecture.} Here we provide details of our network architectures used in the joint upsampling experiments. Our networks have three branches: Encoder, Guidance, and Decoder. The layers in each branch of the joint depth upsampling networks are listed in Tab.~\ref{tab:upsample_arch}. Since we use each PAC$^\intercal$ for 2$\times$ upsampling, 4$\times$, 8$\times$, 16$\times$ networks requires 2, 3, 4 PAC$^\intercal$ layers respectively. The final output from the guidance branch is equally divided in the channel dimension for use as adapting features for the PAC$^\intercal$ layers in the decoder. All CONV and PAC$^\intercal$ layers use $5\times5$ filters, and are followed by ReLU except for the last CONV. We use Gaussian kernels for $K$ in all PAC$^\intercal$ layers. 

We design two variants of our model, \emph{standard} and \emph{lite}. The \emph{standard} variant has a simpler design, but has varying number of parameters for different upsampling factors, and overall consume more memory 
than DJF~\cite{li2016deep}, a previous state-of-the-art approach on joint depth upsampling. For the \emph{lite} variant, we reduce the number of filters and make sure the networks roughly match the number of parameters compared to DJF. 

Similar network architectures are also used for optical flow upsampling. First layer of encoder and last layer in  decoder are modified to fit the two $(u,v)$ channels in optical flow instead of one channel in depth maps, \ie using ``C2" instead of ``C1" in Tab.~\ref{tab:upsample_arch}.

\begin{table}[h!]
    \small
    \centering
    \begin{tabular}{r|c|c|c|c|c|c}
\toprule
    	& 	\multicolumn{3}{c|}{standard} & 	\multicolumn{3}{c}{lite}		\\
\midrule
	& 	4$\times$&	8$\times$& 	16$\times$& 4$\times$& 	8$\times$& 16$\times$	\\
\midrule
\multirow{3}{*}{Encoder}	& 	C32	& 	C32	& 	C32	& 	C12	& 	C12	& 	C8	\\
	& 	C32	& 	C32	& 	C32	& 	C16	& 	C16	& 	C16	\\
	& 	C32	& 	C32	& 	C32	& 	C22	& 	C16	& 	C16	\\
\midrule
\multirow{3}{*}{Guidance}	& 	C32	& 	C32	& 	C32	& 	C12	& 	C12	& 	C8	\\
	& 	C32	& 	C32	& 	C32	& 	C22	& 	C16	& 	C16	\\
	& 	C32	& 	C48	& 	C64	& 	C24	& 	C36	& 	C40	\\
\midrule
\multirow{6}{*}{Decoder}	& 	P32	& 	P32	& 	P32	& 	P12	& 	P12	& 	P8	\\
	& 	P32	& 	P32	& 	P32	& 	P16	& 	P16	& 	P16	\\
	& 	C32	& 	P32	& 	P32	& 	C22	& 	P16	& 	P16	\\
	& 	C1	& 	C32	& 	P32	& 	C1	& 	C20	& 	P16	\\
	& 		& 	C1	& 	C32	& 		& 	C1	& 	C16	\\
	& 		& 		& 	C1	& 		& 		& 	C1	\\
\midrule
\#Params	& 	183K	& 	222K	& 	260K	& 	56K	& 	56K	& 	56K \\
\bottomrule
    \end{tabular}
    \vspace{1.5mm}
    \caption{\textbf{Network architectures for joint depth upsampling.} ``C" stands for regular CONV, ``P" stands for PAC$^\intercal$ (the transposed convolution variant of PAC), and the number after them represents the number of output channels.}
    \label{tab:upsample_arch}
\end{table}

\paragraph{Additional examples.} We provide more joint upsampling visual results for depth (Fig.~\ref{fig:upsample_depth_more}) and optical flow (Fig.~\ref{fig:upsample_flow_more}). 

\section{Conditional Random Fields}
\label{sec:supp_crf}

\paragraph{Interpretations of the formulation.}
The pairwise potentials in Full-CRF is defined as $\psi_p(l_i, l_j|I) = \mu(l_i, l_j)K(\mathbf{f}_i, \mathbf{f}_j) $, where the kernel function $K$ has two terms, \emph{appearance kernel} and \emph{smoothness kernel}

\begin{align}
    K(\mathbf{f}_i, \mathbf{f}_j) = &w_1 \underbrace{\exp \biggr\{-\frac{\|\mathbf{p}_i - \mathbf{p}_j\|^2}{2\theta_\alpha^2}-\frac{\|I_i-I_j\|^2}{2\theta_\beta^2}\biggr\}}_\text{appearance kernel} \nonumber \\ 
    &+ w_2 \underbrace{\exp \biggr\{ -\frac{\|\mathbf{p}_i - \mathbf{p}_j\|^2}{2\theta_\gamma^2} \biggr\}}_\text{smoothness kernel} \label{eq:k_fullcrf_supp}
\end{align}

In comparision, our pairwise potential uses (assuming using Guassian kernel and a single pairwise term)

\begin{align}
    K'(\mathbf{f}_i,\mathbf{f}_j) &= \mathbf{W}[\mathbf{p}_j-\mathbf{p}_i] K(\mathbf{f}_i,\mathbf{f}_j) \nonumber \\
    &=\mathbf{W}[\mathbf{p}_j-\mathbf{p}_i] \exp \left\{-\frac{1}{2} \|\mathbf{f}_i - \mathbf{f}_j\|^2 \right\} \label{eq:k_paccrf_2d}
\end{align}

There are two major differences: 
\begin{enumerate}
    \item The \emph{smoothness kernel} is now moved out of $K$ and is represented using filter $\mathbf{W}$. It can still be initialized as a Gaussian, but arbitrary filter is allowed to be learned.
    \item The \emph{appearance kernel} now operates on $\mathbf{f}$ directly without the need of decomposing it into multiple parts, and without the individual scaling factors ($\theta_\alpha, \dots$). 
\end{enumerate}

Both changes give the pairwise potential more learning capacity. Note that $\mathbf{f}$ can be the output of some other network layers. A simple linear layer can learn appropriate scaling factors, while in other cases a more complex network may be preferred. For input with more than RGB channels (\eg 3D data with color, depth, normal, curvature, \etc), hand-crafting and finding parameters for kernel functions like Eq.~\ref{eq:k_fullcrf_supp} can be time-consuming and suboptimal, and allowing the function to be learned from data in an end-to-end fashion is particularly desirable.  

Note that in Eq.~\ref{eq:k_paccrf_2d}, $\mathbf{W}$ is a 2D matrix, and the corresponding pairwise potential is defined as 
\begin{align}
    \psi_p(l_i, l_j) = \mu(l_i, l_j) \mathbf{W}[\mathbf{p}_j-\mathbf{p}_i] K(\mathbf{f}_i,\mathbf{f}_j) \label{eq:pairwise_potential_2d}
\end{align}

where $\mu(l_i, l_j)$ is the compatibility matrix.
Our final pairwise potential,  $\psi_p(l_i, l_j) = K(\mathbf{f}_i,\mathbf{f}_j) \mathbf{W}_{l_j l_i}[\mathbf{p}_j-\mathbf{p}_i]$ , can be seen as a further step of generalization, where $\mathbf{W}$ is now a 4D tensor. Intuitively, this formulation allows the label compatibility pattern to be spatially varying across different pixel locations.  Eq.~\ref{eq:pairwise_potential_2d} can be seen as a special case factorizing the 4D tensor as the product of two 2D matrices.

\paragraph{Mean-field inference derivation.}

We will start from the mean-field update equation for general pairwise CRFs, Eq.~\ref{eq:update_kf}. Detailed derivation for it can be found in Koller and Friedman~\cite[Chapter 11.5]{koller2009pgm}. 

\begin{align}
    Q_i(l) = \frac{1}{Z_i} \exp \biggr\{ -\psi_u (l) - \sum_{j\in \Omega(i)} \mathbf{E}_{l_j\sim Q_j } \psi_p(l,l_j) \biggr\} \label{eq:update_kf}
\end{align}

Considering that we use multiple neighborhoods (with different dilation factors) in parallel, the update equation becomes

\begin{align}
    Q_i(l) = \frac{1}{Z_i} \exp \biggr\{ -\psi_u (l) - \sum_k\sum_{j\in \Omega^k(i)} \mathbf{E}_{l_j\sim Q_j } \psi_p^k(l,l_j) \biggr\}  \label{eq:update_kf_parallel}
\end{align}

Substituting the pairwise potential with
\begin{align}
    \psi_p^k(l_i, l_j) = K^k(\mathbf{f}_i,\mathbf{f}_j) \mathbf{W}_{l_j l_i}^k[\mathbf{p}_j-\mathbf{p}_i] \label{eq:pairwise_paccrf_parallel}
\end{align}

the update rule becomes
\vspace{-4mm}
\begin{align}
    Q_i(l) = &\frac{1}{Z_i} \exp \biggr\{ -\psi_u (l) \nonumber \\
    &- \sum_k \sum_{j\in \Omega^k(i)} \mathbf{E}_{l_j\sim Q_j} \biggr\{K^k(\mathbf{f}_i,\mathbf{f}_j) \mathbf{W}_{l_j l}^k[\mathbf{p}_j-\mathbf{p}_i] \biggr\} \biggr\} \nonumber \\
    = &\frac{1}{Z_i} \exp \biggr\{ -\psi_u (l) \nonumber\\ 
    &- \sum_k \sum_{l' \in \mathcal{L}}\sum_{j\in \Omega^k(i)}K^k(\mathbf{f}_i, \mathbf{f}_j) \mathbf{W}^k_{l'l}[\mathbf{p}_j-\mathbf{p}_i] Q_j(l') \biggr\} \label{eq:update_q} 
\end{align}

Using Eq.~\ref{eq:update_q} in an iterative fashion leads to the final update rule of mean-field inference: 

\begin{align}
    &Q_i^{(t+1)}(l) \leftarrow \frac{1}{Z_i} \exp \biggr\{ -\psi_u (l) \nonumber\\ &- \sum_k\underbrace{\sum_{l' \in \mathcal{L}}\sum_{j\in \Omega^k(i)}K^k(\mathbf{f}_i, \mathbf{f}_j) \mathbf{W}^k_{l'l}[\mathbf{p}_j-\mathbf{p}_i] Q_j^{(t)}(l')}_\text{PAC} \biggr\} \label{eq:update_paccrf_supp}
\end{align}

\paragraph{Mean-field inference steps.}

Tab.~\ref{tab:crf_mfsteps} shows how mIoU changes with different mean-field steps. We use 5 steps for all other experiments in the paper.

\begin{table}[ht!]
    \centering
    \mycaption{Impact of MF steps in PAC-CRF}{Validation mIoU when using different number of MF steps in PAC-CRF.}
    \small
    \begin{tabular}{lcccc}
    \toprule
    Mean-field steps  & 1 & 3 & 5 & 7 \\
    \midrule
    mIoU & 68.38  & 68.72 & \textbf{68.90} & 68.90 \\
    time & 19 ms & 49 ms & 78 ms & 109 ms \\
    \bottomrule
    \end{tabular}
    \label{tab:crf_mfsteps}
\end{table}

\paragraph{On the contribution of dilation.}
Just like standard convolution, PAC supports dilation to increase the receptive field without increasing the number of parameters. This capability is leveraged by PAC-CRF to allow long-range connections. For a similar purpose, Conv-CRF applies Gaussian blur to pairwise potentials to increase the receptive field. 
To quantify the improvements due to dilation, we try another baseline where we add dilation to Conv-CRF. The improved performance (+2.13/+1.57 \(\rightarrow\) +2.50/+1.91) validates that dilation is indeed an important ingredient, while the remaining gap shows that the PAC formulation is essential to the full gain. 

\renewcommand{\factor}{0.16}
\begin{figure*}
\captionsetup[subfigure]{labelformat=empty}
\begin{centering}

\subfloat{\includegraphics[width=\factor\textwidth]{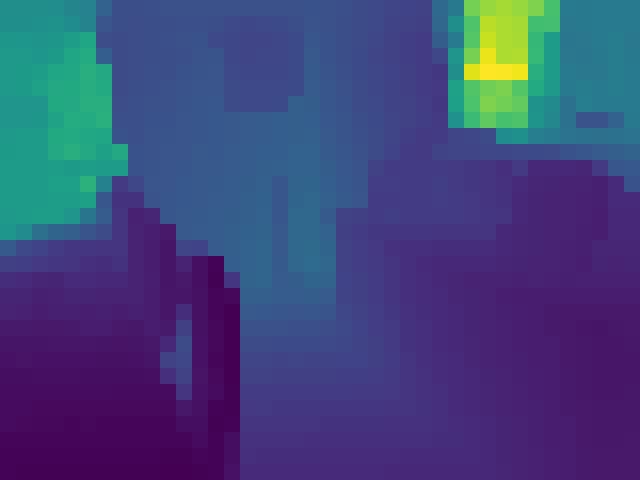}}~
\subfloat{\includegraphics[width=\factor\textwidth]{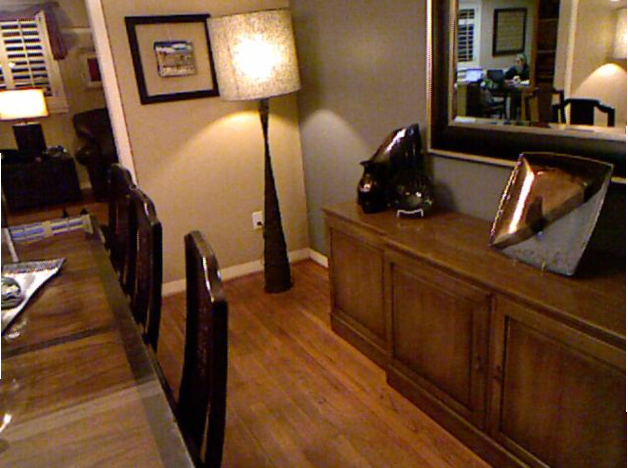}}~
\subfloat{\includegraphics[width=\factor\textwidth]{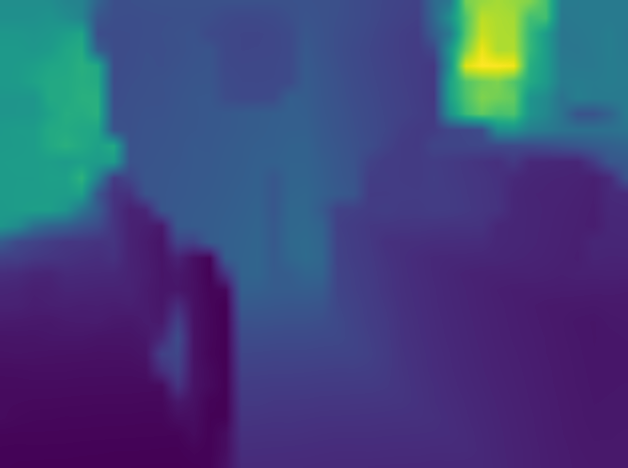}}~
\subfloat{\includegraphics[width=\factor\textwidth]{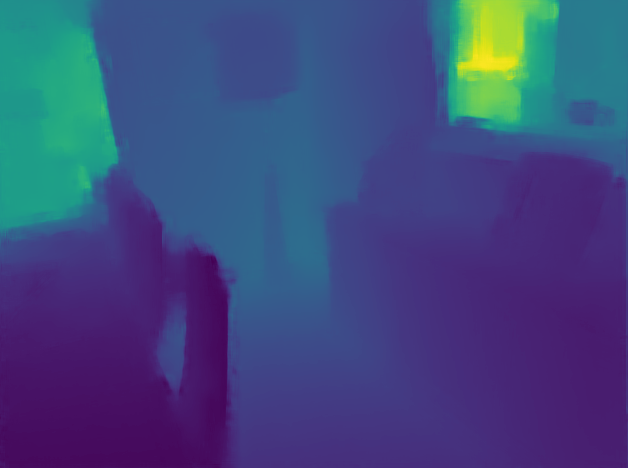}}~
\subfloat{\includegraphics[width=\factor\textwidth]{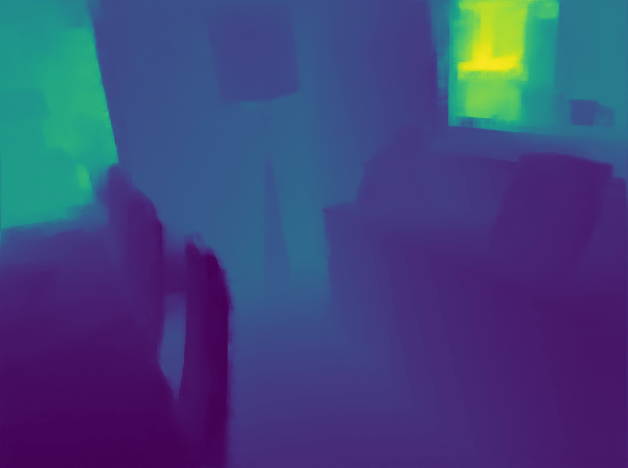}}~
\subfloat{\includegraphics[width=\factor\textwidth]{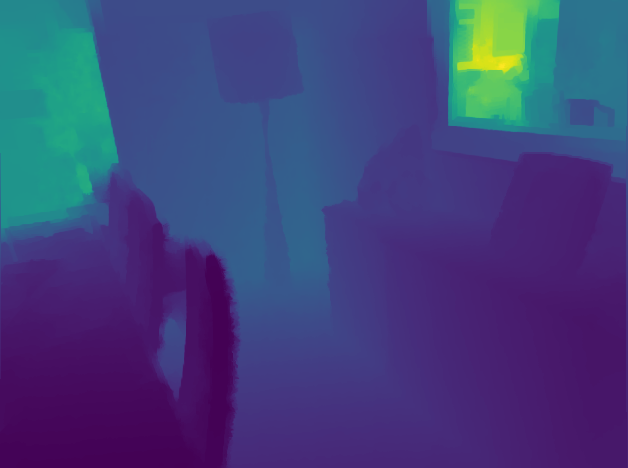}}~\\
\vspace{-2mm}

\subfloat{\includegraphics[width=\factor\textwidth]{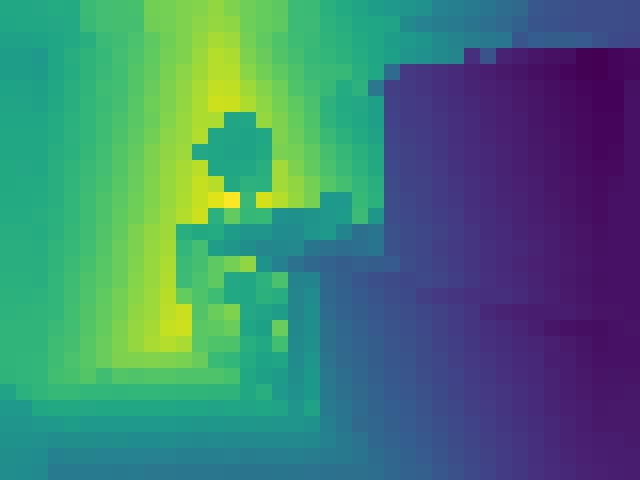}}~
\subfloat{\includegraphics[width=\factor\textwidth]{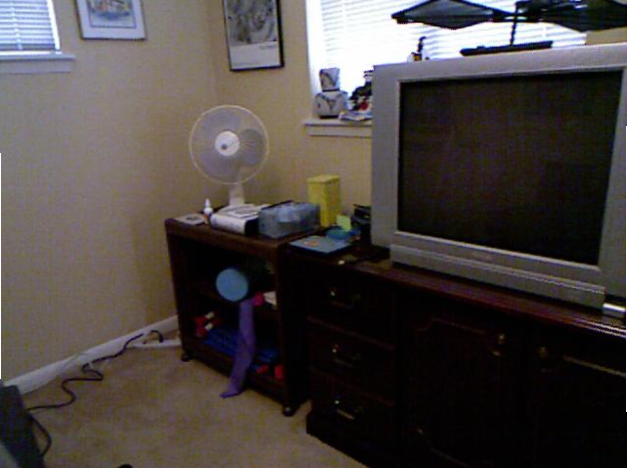}}~
\subfloat{\includegraphics[width=\factor\textwidth]{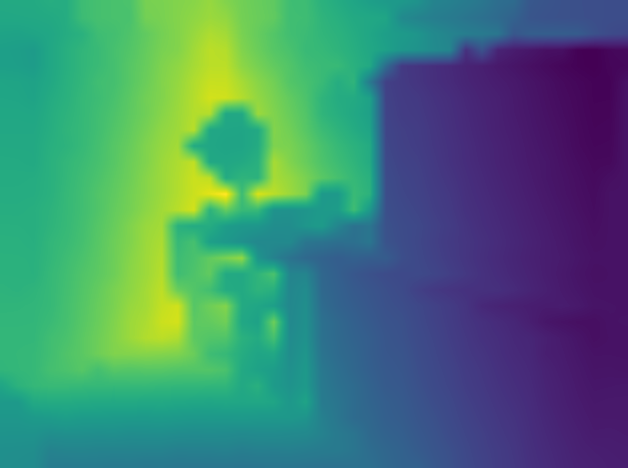}}~
\subfloat{\includegraphics[width=\factor\textwidth]{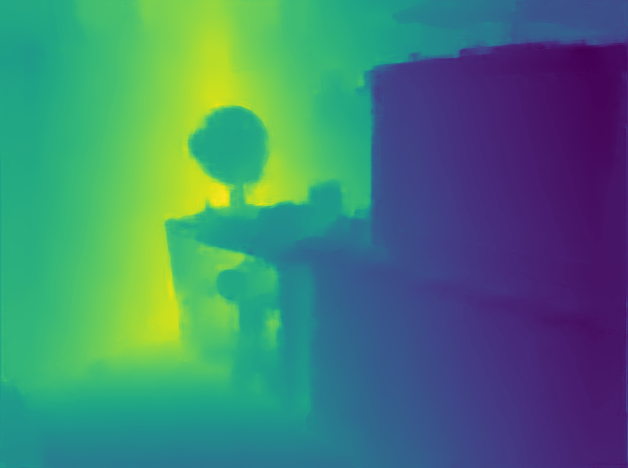}}~
\subfloat{\includegraphics[width=\factor\textwidth]{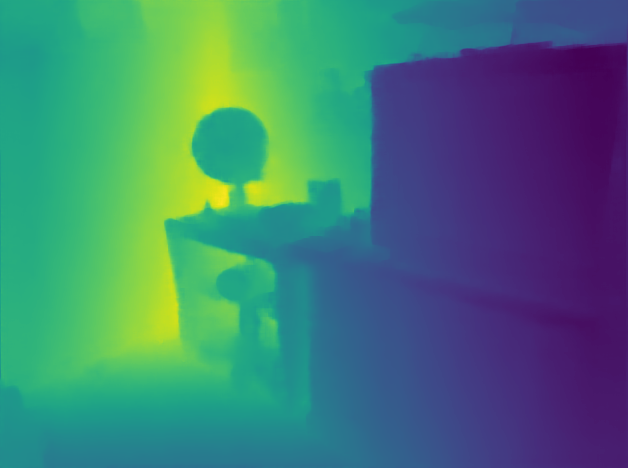}}~
\subfloat{\includegraphics[width=\factor\textwidth]{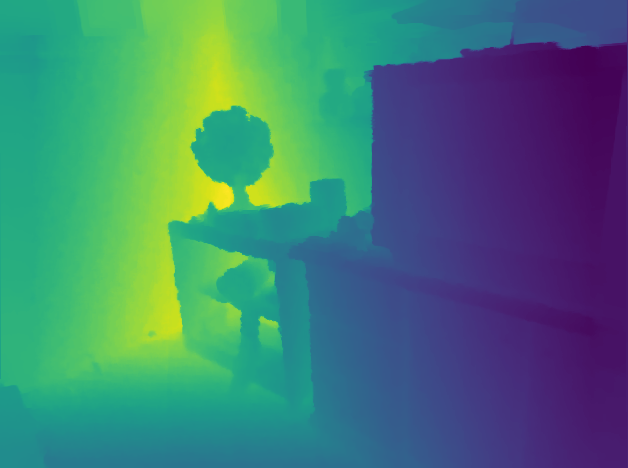}}~\\
\vspace{-2mm}

\subfloat{\includegraphics[width=\factor\textwidth]{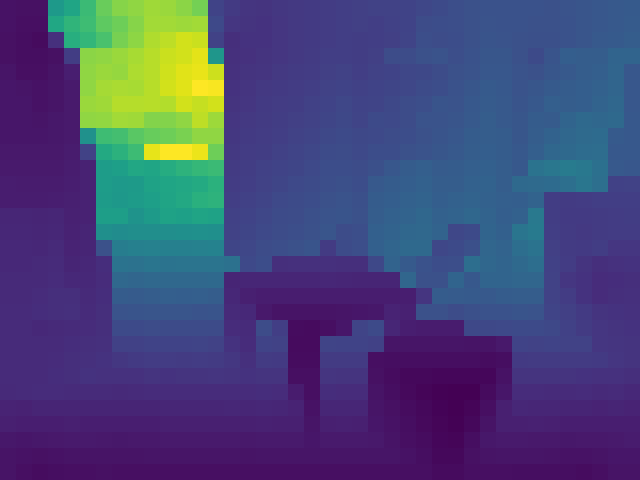}}~
\subfloat{\includegraphics[width=\factor\textwidth]{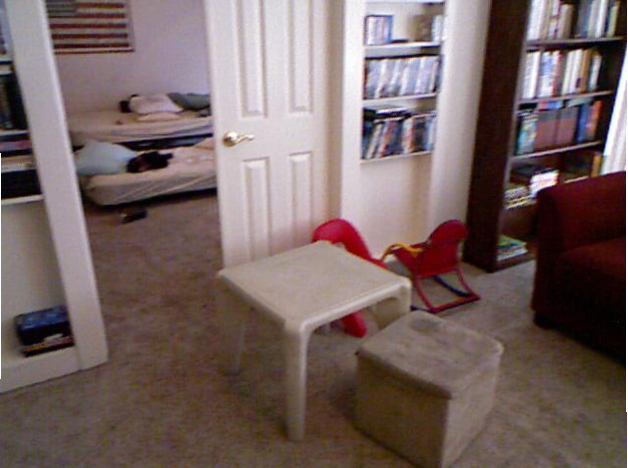}}~
\subfloat{\includegraphics[width=\factor\textwidth]{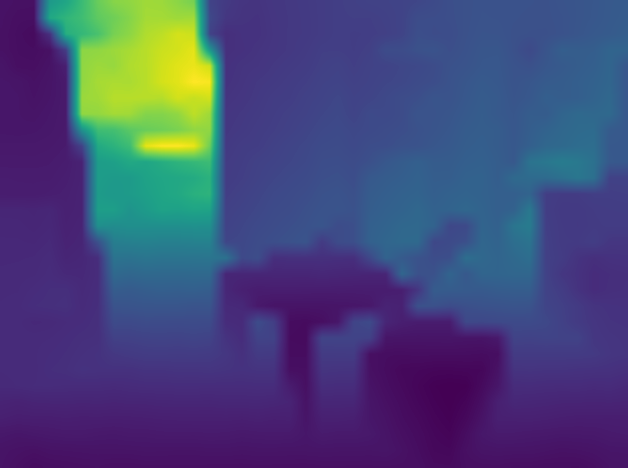}}~
\subfloat{\includegraphics[width=\factor\textwidth]{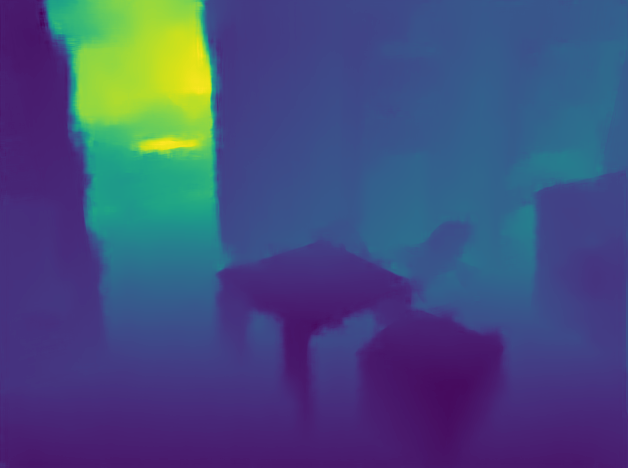}}~
\subfloat{\includegraphics[width=\factor\textwidth]{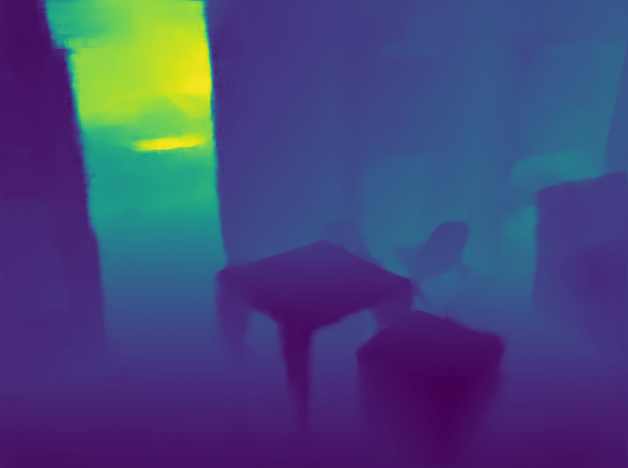}}~
\subfloat{\includegraphics[width=\factor\textwidth]{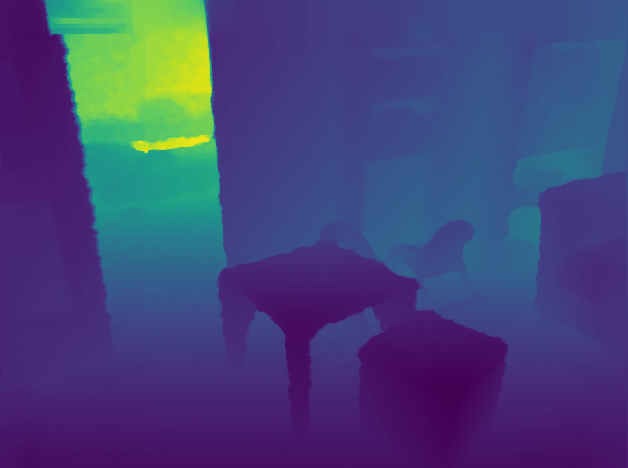}}~\\
\vspace{-2mm}

\subfloat{\includegraphics[width=\factor\textwidth]{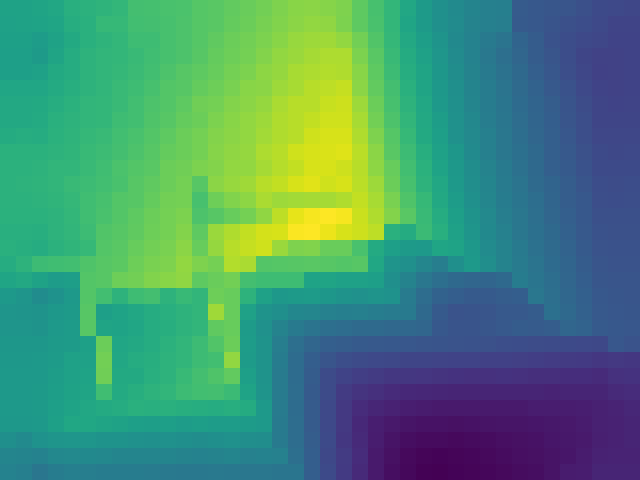}}~
\subfloat{\includegraphics[width=\factor\textwidth]{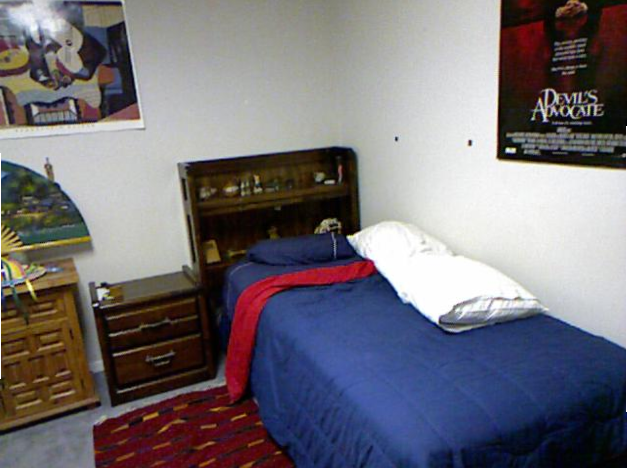}}~
\subfloat{\includegraphics[width=\factor\textwidth]{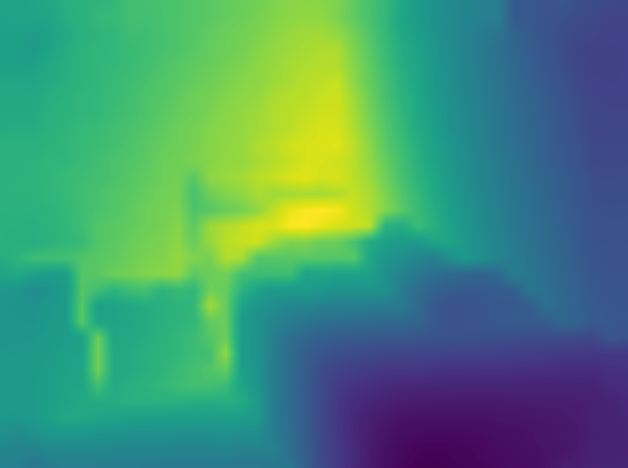}}~
\subfloat{\includegraphics[width=\factor\textwidth]{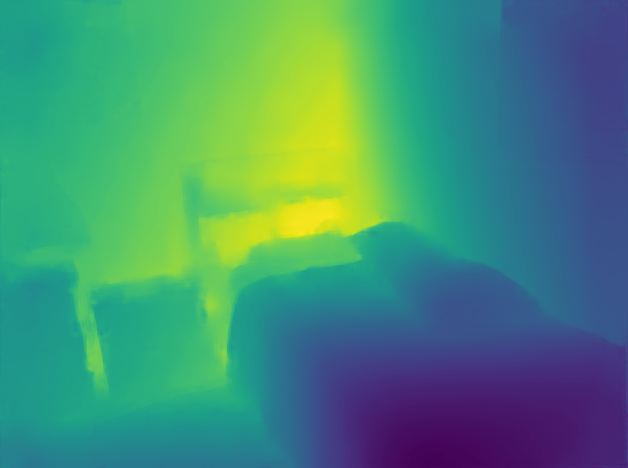}}~
\subfloat{\includegraphics[width=\factor\textwidth]{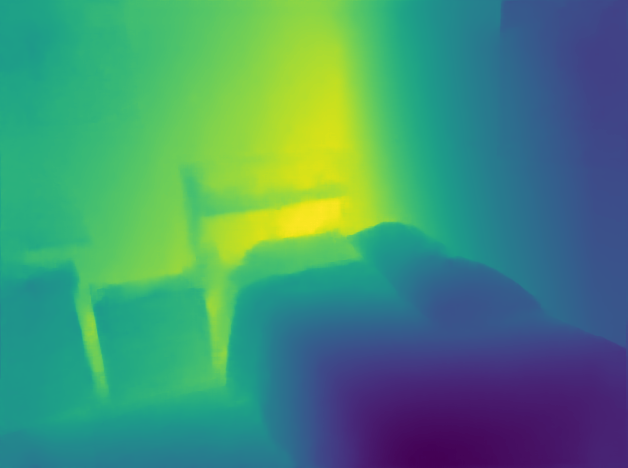}}~
\subfloat{\includegraphics[width=\factor\textwidth]{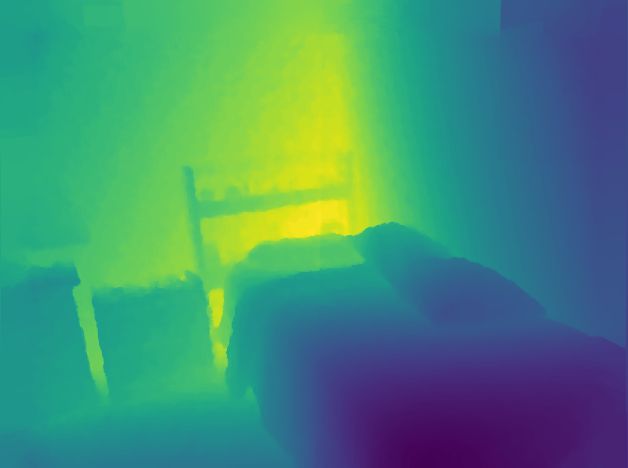}}~\\
\vspace{-2mm}

\setcounter{subfigure}{0}
\subfloat[Input ($1/16$ res.)]{\includegraphics[width=\factor\textwidth]{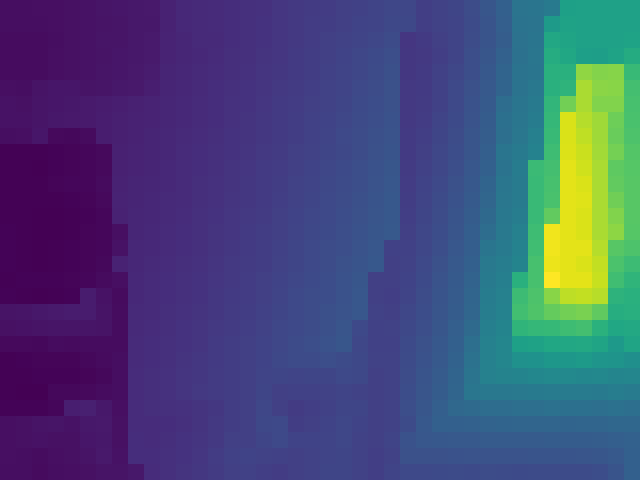}}~
\subfloat[Guide]{\includegraphics[width=\factor\textwidth]{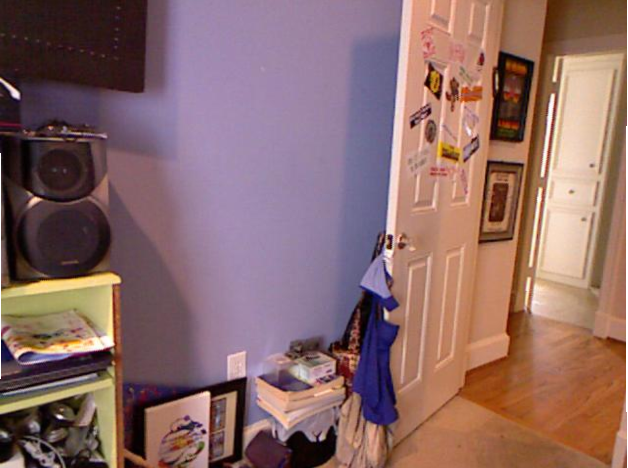}}~
\subfloat[Bilinear]{\includegraphics[width=\factor\textwidth]{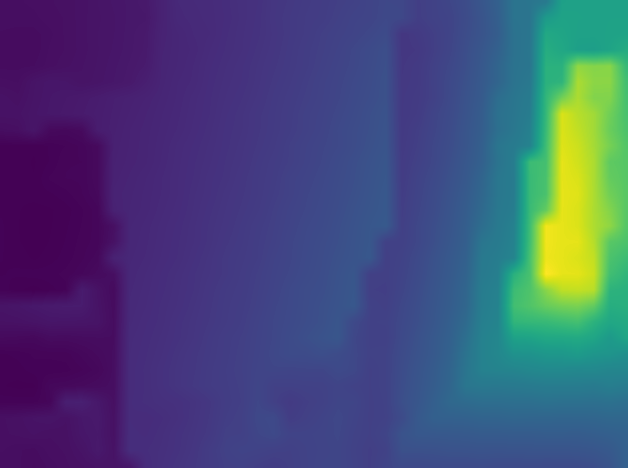}}~
\subfloat[DJF~\cite{li2016deep}]{\includegraphics[width=\factor\textwidth]{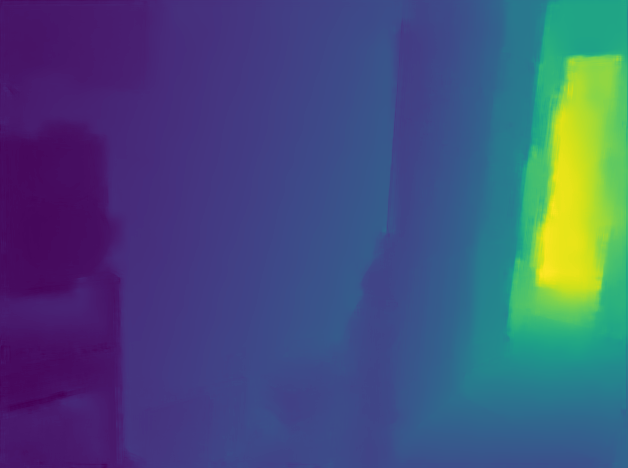}}~
\subfloat[Ours]{\includegraphics[width=\factor\textwidth]{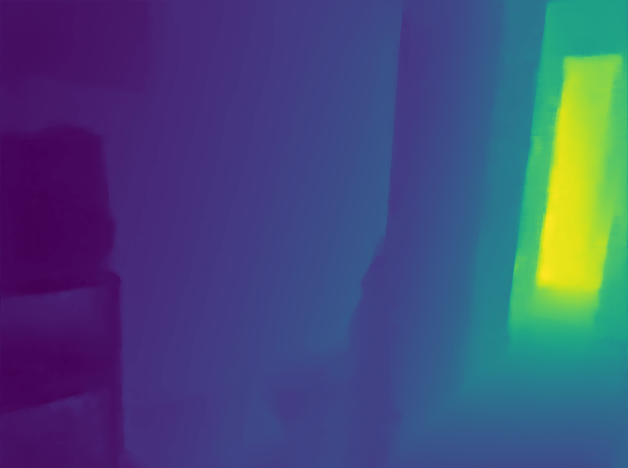}}~
\subfloat[GT]{\includegraphics[width=\factor\textwidth]{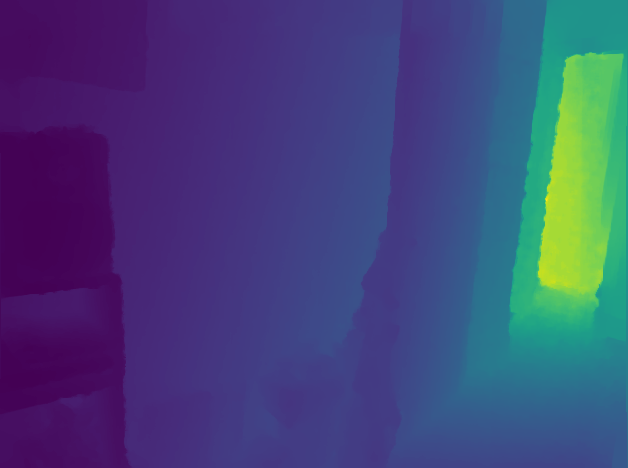}}~\\
\vspace{2mm}
\mycaption{Additional examples of joint depth upsampling}{Samples are from the test set of NYU Depth V2. Zoom in for full details.}\label{fig:upsample_depth_more}
\end{centering}
\end{figure*}

\begin{figure*}
\captionsetup[subfigure]{labelformat=empty}
\begin{centering}

\subfloat{\includegraphics[width=\factor\textwidth]{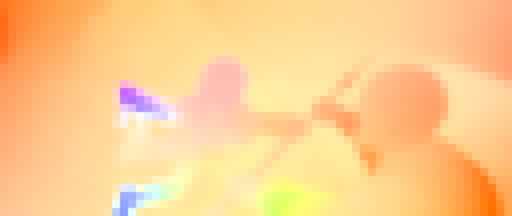}}~
\subfloat{\includegraphics[width=\factor\textwidth]{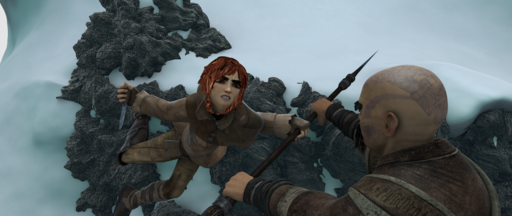}}~
\subfloat{\includegraphics[width=\factor\textwidth]{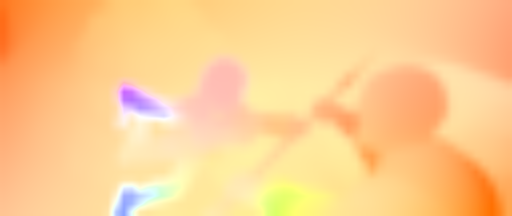}}~
\subfloat{\includegraphics[width=\factor\textwidth]{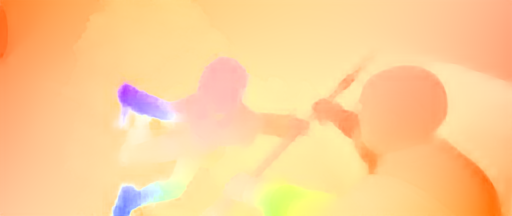}}~
\subfloat{\includegraphics[width=\factor\textwidth]{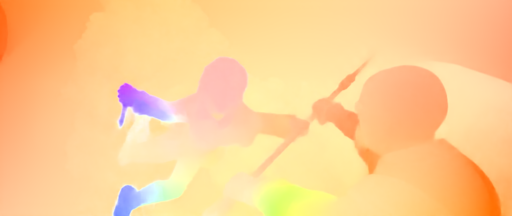}}~
\subfloat{\includegraphics[width=\factor\textwidth]{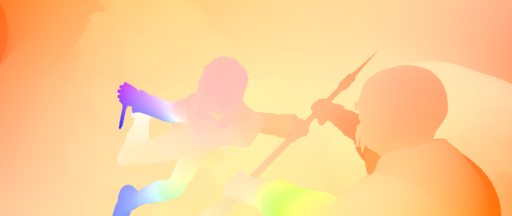}}~\\
\vspace{-2mm}

\subfloat{\includegraphics[width=\factor\textwidth]{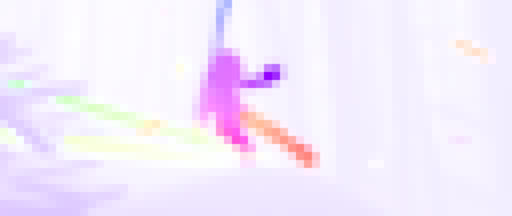}}~
\subfloat{\includegraphics[width=\factor\textwidth]{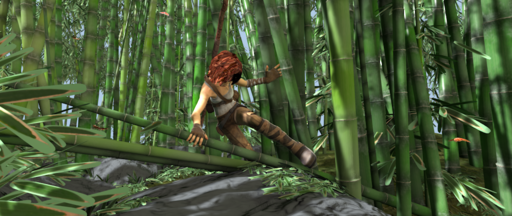}}~
\subfloat{\includegraphics[width=\factor\textwidth]{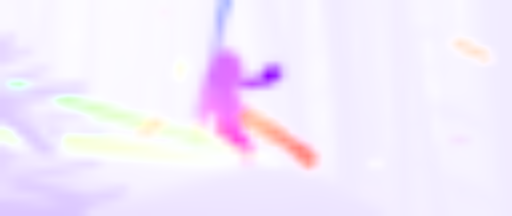}}~
\subfloat{\includegraphics[width=\factor\textwidth]{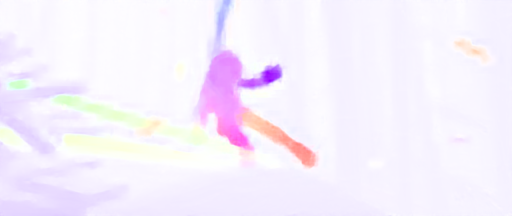}}~
\subfloat{\includegraphics[width=\factor\textwidth]{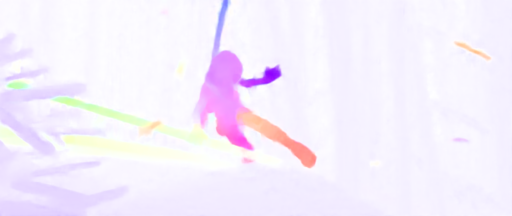}}~
\subfloat{\includegraphics[width=\factor\textwidth]{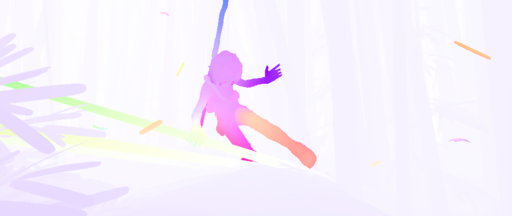}}~\\
\vspace{-2mm}

\subfloat{\includegraphics[width=\factor\textwidth]{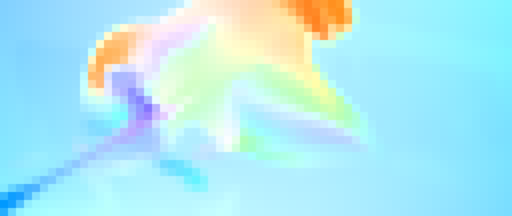}}~
\subfloat{\includegraphics[width=\factor\textwidth]{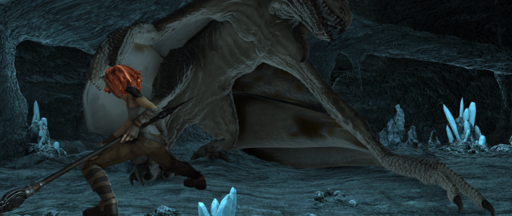}}~
\subfloat{\includegraphics[width=\factor\textwidth]{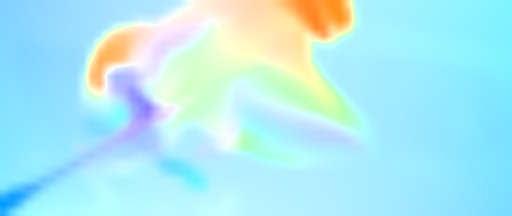}}~
\subfloat{\includegraphics[width=\factor\textwidth]{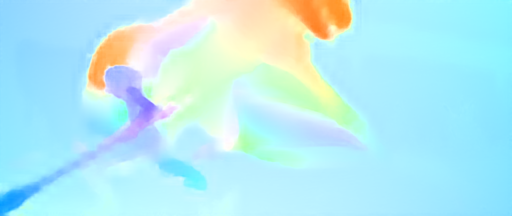}}~
\subfloat{\includegraphics[width=\factor\textwidth]{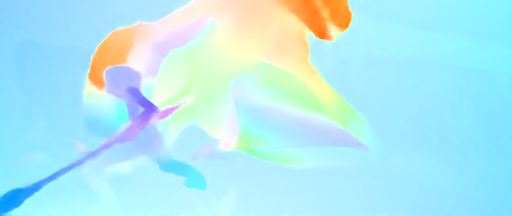}}~
\subfloat{\includegraphics[width=\factor\textwidth]{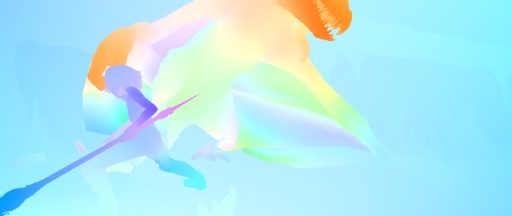}}~\\
\vspace{-2mm}

\subfloat{\includegraphics[width=\factor\textwidth]{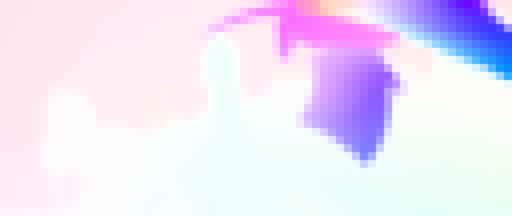}}~
\subfloat{\includegraphics[width=\factor\textwidth]{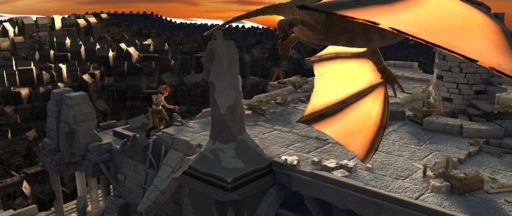}}~
\subfloat{\includegraphics[width=\factor\textwidth]{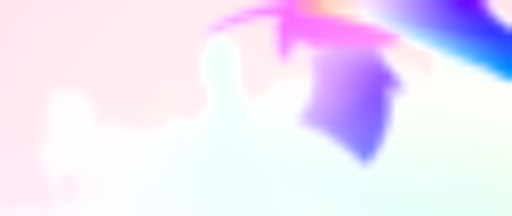}}~
\subfloat{\includegraphics[width=\factor\textwidth]{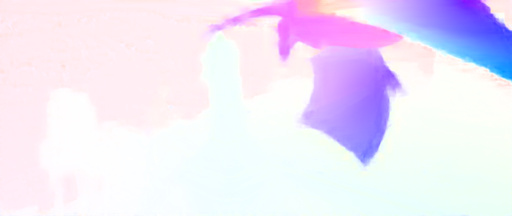}}~
\subfloat{\includegraphics[width=\factor\textwidth]{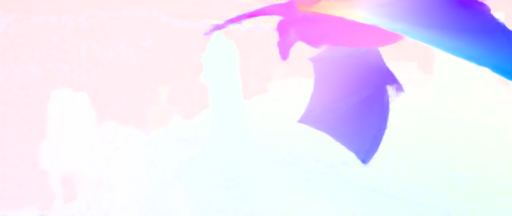}}~
\subfloat{\includegraphics[width=\factor\textwidth]{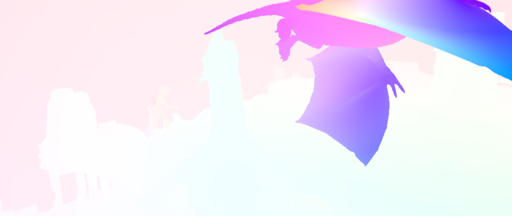}}~\\
\vspace{-2mm}

\subfloat{\includegraphics[width=\factor\textwidth]{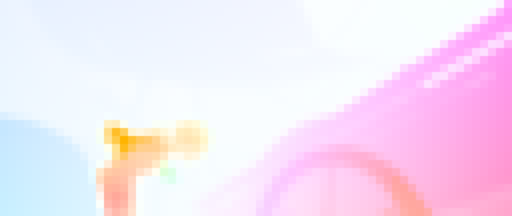}}~
\subfloat{\includegraphics[width=\factor\textwidth]{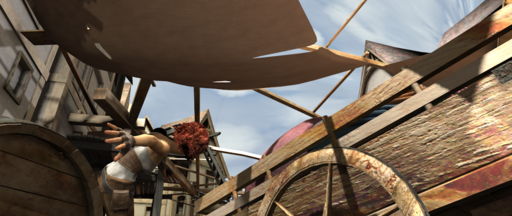}}~
\subfloat{\includegraphics[width=\factor\textwidth]{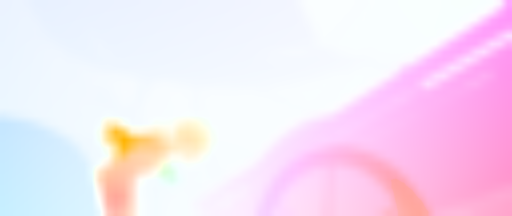}}~
\subfloat{\includegraphics[width=\factor\textwidth]{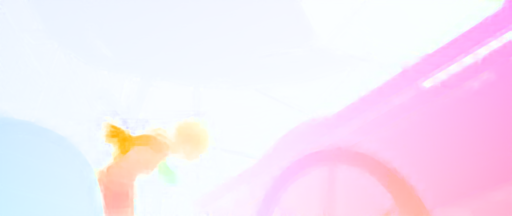}}~
\subfloat{\includegraphics[width=\factor\textwidth]{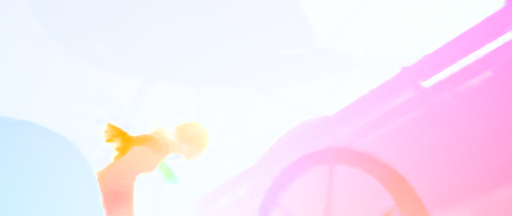}}~
\subfloat{\includegraphics[width=\factor\textwidth]{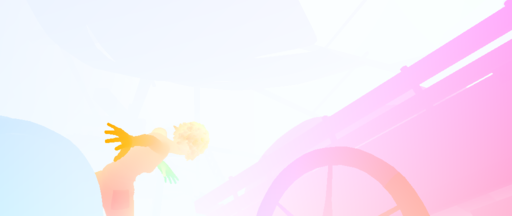}}~\\
\vspace{-2mm}

\setcounter{subfigure}{0}
\subfloat[Input ($1/16$ res.)]{\includegraphics[width=\factor\textwidth]{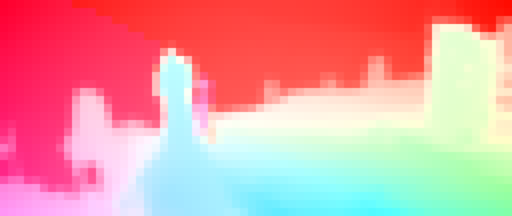}}~
\subfloat[Guide (1\textsuperscript{st} frame)]{\includegraphics[width=\factor\textwidth]{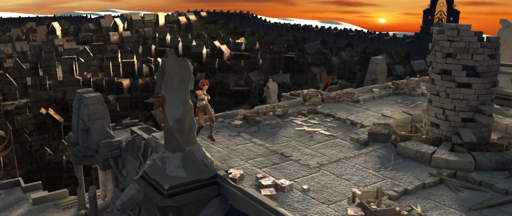}}~
\subfloat[Bilinear]{\includegraphics[width=\factor\textwidth]{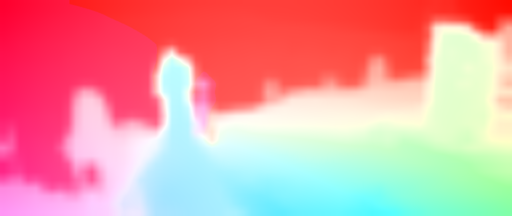}}~
\subfloat[DJF~\cite{li2016deep}]{\includegraphics[width=\factor\textwidth]{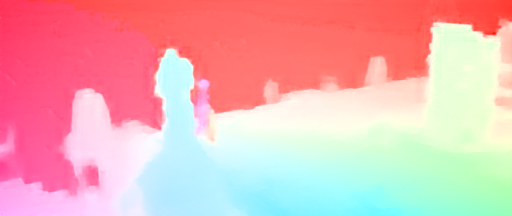}}~
\subfloat[Ours]{\includegraphics[width=\factor\textwidth]{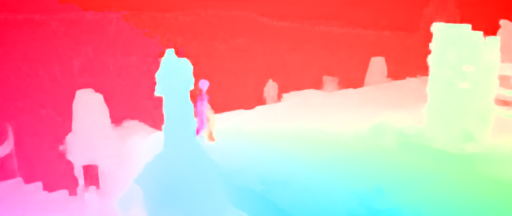}}~
\subfloat[GT]{\includegraphics[width=\factor\textwidth]{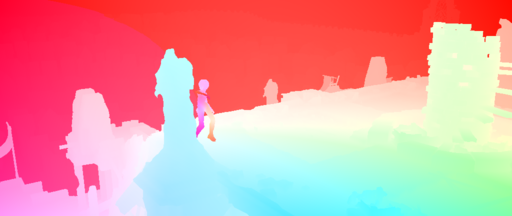}}~\\
\vspace{2mm}
\mycaption{Additional examples of joint optical flow upsampling}{Samples are from the val set of Sintel. Zoom in for full details. }\label{fig:upsample_flow_more}
\end{centering}
\end{figure*}